\documentclass[preprint,12pt]{larticle}
\usepackage[a4paper, total={6.18in, 10in}]{geometry}
\usepackage[utf8]{inputenc}

\usepackage{amssymb}
\usepackage{amsthm}
\usepackage{amsmath}

\usepackage{CJKutf8}
\usepackage{tikz}
\usepackage{tikz-qtree}

\usepackage[symbol]{footmisc}

\newcommand{\M}{\mathcal{V}}
\newcommand{\abs}[1]{\left\vert#1\right\vert}

\usepackage{xcolor}

\begin{document}

\begin{frontmatter}

\title{Quantifying syntax similarity with a polynomial representation of dependency trees}
%\title{Polynomial representation and comparison of dependency trees}

\author{Pengyu Liu\footnotesize{$^{1,2}$}}

\author{Tinghao Feng\footnotesize{$^{3}$}}

\author{Rui Liu\footnotesize{$^{4,5}$}\footnote{To whom correspondence should be addressed; e-mail: \url{liu_rui@bnu.edu.cn}.}}

\address{\footnotesize{$^1$}Department of Microbiology and Molecular Genetics, University of California, Davis, Davis, CA 95616, USA}
\address{\footnotesize{$^2$}Department of Mathematics, Simon Fraser University, Burnaby, BC V5A 1S6, Canada}
\address{\footnotesize{$^3$}Department of Computer Science, Appalachian State University, Boone, NC 28608, USA}
\address{\footnotesize{$^4$}Department of Chinese Language and Literature and \footnotesize{$^5$}Center for Linguistic Sciences, Beijing Normal University, Zhuhai, Guangdong, China}

\begin{abstract}
We introduce a graph polynomial that distinguishes tree structures to represent dependency grammar and a measure based on the polynomial representation to quantify syntax similarity.
The polynomial encodes accurate and comprehensive information about the dependency structure and dependency relations of words in a sentence.
We apply the polynomial-based methods to analyze sentences in the Parallel Universal Dependencies treebanks.
Specifically, we compare the syntax of sentences and their translations in different languages, and we perform a syntactic typology study of available languages in the Parallel Universal Dependencies treebanks.
We also demonstrate and discuss the potential of the methods in measuring syntax diversity of corpora. 
\end{abstract}

\end{frontmatter}

% \linenumbers

\section{Introduction}
\label{S:1}

Dependency grammar is an important framework for syntactic analysis \cite{Imrenyi2020}.
Dependency focuses on the proximity of words in a sentence, and the hierarchical relations between words in the sentence are represented by a tree structure called the dependency tree of the sentence.
Recently, an international collaboration project called Universal Dependency (UD) has created a standard annotation scheme for constructing dependency trees from sentences, and hundreds of UD treebanks of various languages have been made publicly available \cite{deMarneffe2021}.
These datasets form key materials for syntax analysis, providing new opportunities for automated text processing and syntactic typology studies to name a few.
Parallel Universal Dependency (PUD) treebanks are a class of UD treebanks consisting of dependency trees of 1,000 sentences and their translations to other languages \cite{Zeman2017} .
The 1,000 sentences are randomly selected from the news domain and Wikipedia and are originally written in English, French, German, Italian or Spanish.
At the time of writing, there are 20 PUD treebanks containing the dependency trees of the 1,000 sentences in 20 languages respectively.
These UD treebanks have stimulated novel computational methods for syntax analysis and the development of quantitative measures for syntax similarity \cite{Liu2012,Vulic2020,Wong2017}. 
However, current methods describing dependency trees mainly focus on partial syntactic information recorded in the structures such as the order of words and the dependency distance \cite{Chen2017,Chen2022,Gerdes2021,Lei2020}.
In this work, we introduce a comprehensive representation of dependency trees based on a tree distinguishing polynomial.
The polynomial takes into account all syntactic information recorded in a dependency tree, and two sentences have the same dependency structure if and only if the polynomials of their dependency trees are identical.

Structural polynomials are well studied objects in mathematical areas such as knot theory and graph theory, and they have natural applications in characterizing topological and discrete structures. In the theory of knots and links, Jones polynomial \cite{Jones1985} and HOMFLY polynomial \cite{HOMFLY1985} have been used to characterize properties of knots and links such as crossing number \cite{Kauffman1987,Thistlethwaite1987} and braid index \cite{Diao2020,Murasugi1991}. 
In the study of graphs, the Tutte polynomial \cite{Tutte1954} contains the information about graphs including the number of spanning trees of the graph and the number of graph colorings. 
Recently, a structural polynomial that distinguishes unlabeled trees has been defined and studied \cite{Liu2021}.
This builds an one-to-one correspondence between unlabeled trees and a class of bivariate polynomials, that is, two unlabeled trees are isomorphic if and only if they have the same polynomial. 
This tree distinguishing polynomial has been applied to study phylogenetic trees and pathogen evolution \cite{Liu2022} and generalized to represent some classes of phylogenetic networks \cite{Janssen2021,Pons2022,vanIersel2022}.
It has been shown that the polynomial-based methods for tree comparison have better accuracy and computational efficiency, when compared to other tree comparison and representation methods such as sequence-based representations, Laplacian spectrum of trees and summary statistics \cite{Liu2022}. 
Current methods to compare dependency trees are mainly based on summary statistics including tree kernels and their generalizations \cite{Culotta2004, Luo2005} or tree edit distances \cite{Reis2004} which only take into account local structures rather than the global structure of trees.
Here, we generalize the tree distinguishing polynomial for representing dependency trees and define a distance between the polynomials to measure syntax similarity.
We apply the polynomial-based methods to the dependency trees in the PUD treebanks, and we compare the syntax of sentences with small and large distances.
We also perform a syntactic typology study for currently available languages in the PUD treebanks.
Furthermore, we show that the pairwise distances between sentences can be used to measure syntax diversity of a corpus and discuss its potential applications.

\section{Materials and methods}
\label{S:2}

\subsection{Dependency trees}

\begin{figure}[!ht]
\begin{center}
\includegraphics[scale=0.9]{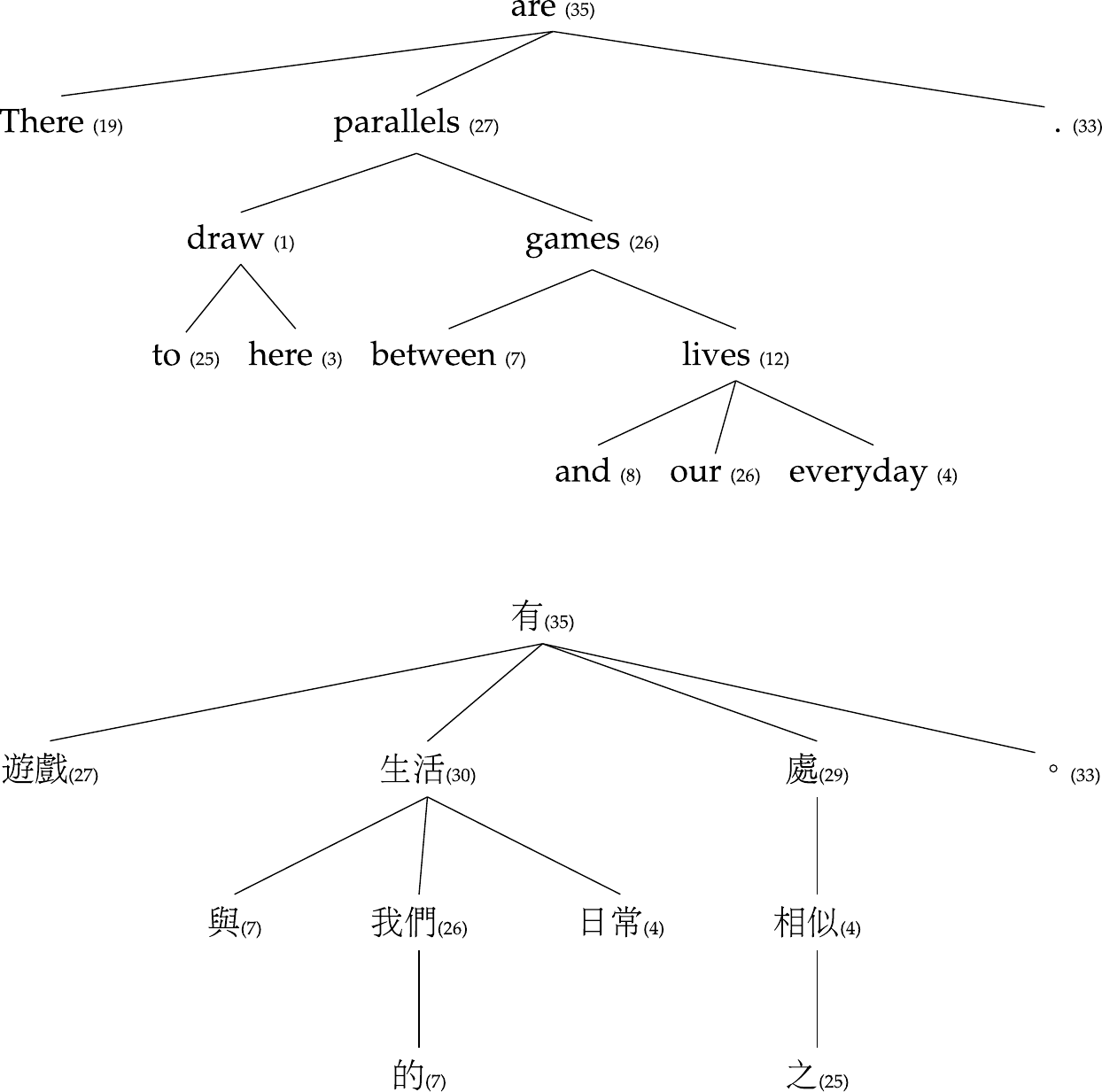}
\renewcommand{\figurename}{Figure}
\caption{{\bf Examples of dependency trees.} Top: the dependency tree of an English sentence: ``There are parallels to draw here between games and our everyday lives.''
Bottom: the dependency tree of a Chinese translation of the sentence. 
The numbers in parentheses after each word are labels representing head-dependent relations listed in Table~\ref{t1}. }\label{f1}
\end{center}
\end{figure}

A {\em dependency tree} of a sentence is a rooted node-labeled tree representing grammatical relations between words in the sentence.
Each node in a dependency tree corresponds to a word in the sentence.
An edge in a dependency tree connects two nodes and represents a grammatical connection between the two corresponding words: The node closer to the root is the {\em head} of the edge and the other node is a {\em dependent} of the head.
A head can have multiple dependents, while every dependent has only one head.
The label of a dependent indicates the grammatical relation to its head.
In a dependency tree of a sentence, the root node representing the head of the entire sentence is not a dependent, so its label only shows that it is the root.
Furthermore, a sentence can contain words with the same grammatical relation, so dependents in a dependency tree can have identical labels.
In Figure~\ref{f1}, we display the dependency tree of an English sentence and the dependency tree of a Chinese translation of the sentence. 
In these examples of dependency trees, the numbers in parentheses after each word are the node labels representing head-dependent grammatical relations listed in Table~\ref{t1}.
All dependency trees used in the paper are constructed by crosslinguistically consistent morphosyntactic annotation under the Universal Dependencies (UD) framework \cite{deMarneffe2021}.

\begin{table}[ht]
    \centering
    \small
\begin{tabular}{ |p{1.1cm}|p{5.3cm}||p{1.1cm}|p{5.3cm}| }
 \hline
Index & Relation & Index& Relation  \\
 \hline
1 & Adjectival clause modifier & 20 &   Fixed multiword expression\\
2 & Adverbial clause modifier  & 21   & Flat multiword expression\\
3 & Adverbial modifier & 22 &  Goes with\\
4 & Adjectival modifier & 23 &  Indirect object\\
5 & Appositional modifier  & 24 & List\\
6 & Auxiliary  & 25   & Marker\\
7 & Case marking  & 26 & Nominal modifier\\
8 & Coordinating conjunction  & 27 & Nominal subject\\
9 & Clausal complement  & 28 & Numeric modifier\\
10 & Classifier  & 29 & Object\\
11 & Compound  & 30 & Oblique nominal\\
12 & Conjunct  & 31 & Orphan\\
13 & Copula  & 32 & Parataxis\\
14 & Clausal subject  & 33 & Punctuation\\
15 & Unspecified dependency  & 34 & Overridden disfluency\\
16 & Determiner  & 35 & Root \\
17 & Discourse element  & 36 & Vocative\\
18 & Dislocated elements  & 37& Open clausal complement\\
19 &  Expletive &  & \\ \hline

\end{tabular}
    \caption{{\bf The indices of head-dependent relations.} The 37 syntactic relations used in the Universal Dependencies (UD) framework \cite{deMarneffe2021}.}
    \label{t1}
\end{table}

\subsection{Parallel Universal Dependencies}

We analyze dependency trees in the Parallel Universal Dependencies (PUD) treebanks, which were created in a shared task of the Conference on Computational Natural Language Learning (CoNLL 2017) \cite{Zeman2017}.
To construct the PUD treebanks, 1,000 sentences were randomly selected from online news or Wikipedia articles, and there were 750 of the sentences originally in English, 100 in German, 50 in French, 50 in Italian and 50 in Spanish.
Then, the 1,000 sentences were translated by professional translators to other languages.
A PUD treebank contains 1,000 dependency trees of the translated or original sentences in a language.
Currently, there are 20 PUD treebanks available, containing dependency trees of the 1,000 translated or original sentences in 20 languages including Arabic, Chinese, Czech, English, Finnish, French, German, Hindi, Icelandic, Indonesian, Italian, Japanese, Korean, Polish, Portuguese, Russian, Spanish, Swedish, Thai and Turkish.

\subsection{Tree distinguishing polynomial}

\begin{figure}[ht]
\begin{center}
\includegraphics[scale=0.9]{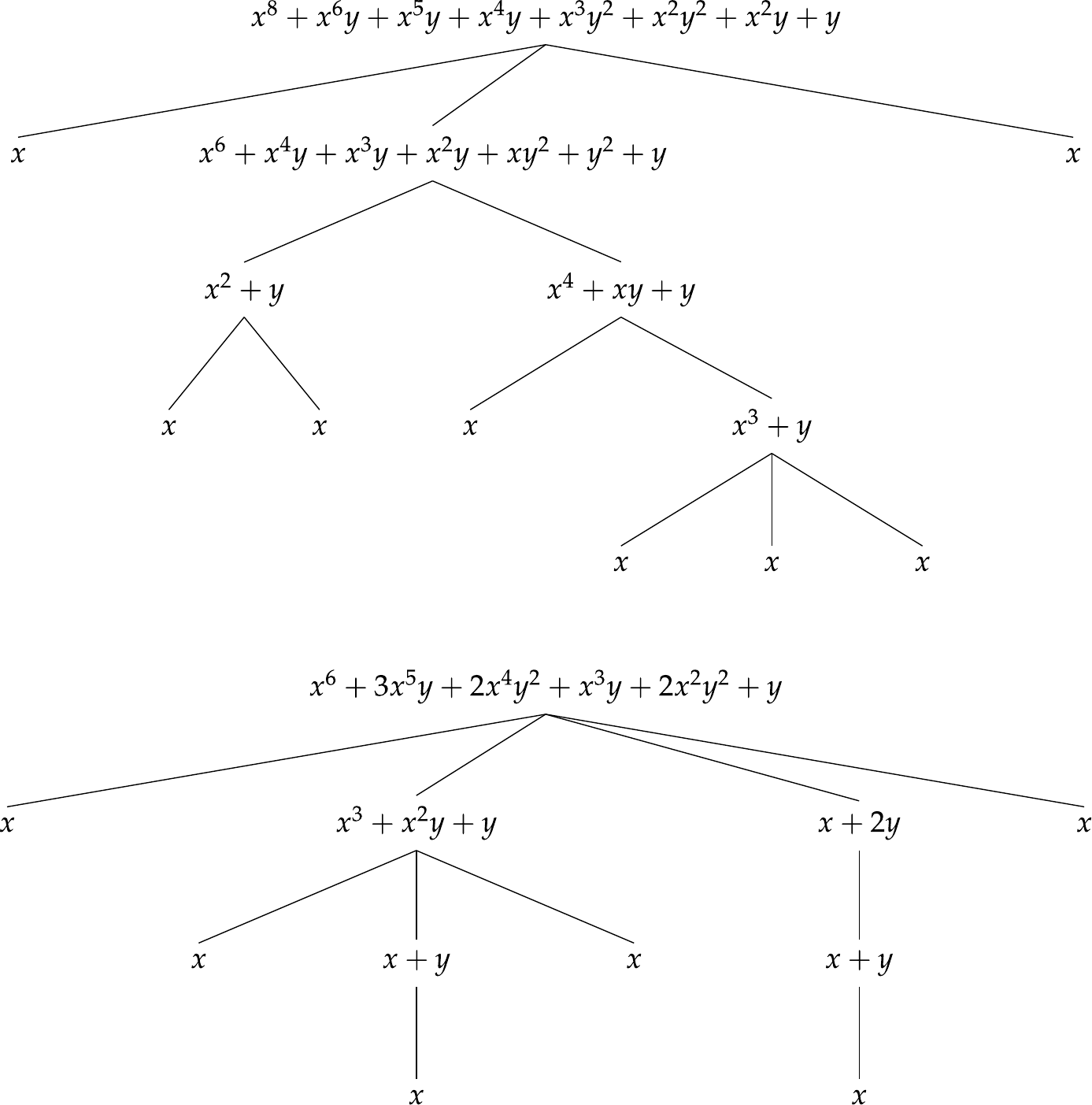}
\renewcommand{\figurename}{Figure}
\caption{{\bf Examples of polynomials of unlabeled trees.} The recursive process of computing the polynomials of topologies of dependency trees displayed in Figure~\ref{f1} from the leaf nodes to the roots.
The polynomials at the roots represent the two unlabeled trees.}\label{f2}
\end{center}
\end{figure}

We review the graph polynomial that distinguishes unlabeled trees introduced in \cite{Liu2021}. 
Every rooted unlabeled tree $T$ corresponds to a unique bivariate polynomial $P(T,x,y)$. 
To compute the polynomial $P(T,x,y)$ for the unlabeled tree $T$, we recursively assign a polynomial to each node in $T$ from the leaf nodes to the root, and the polynomial at the root is $P(T,x,y)$.
Let $P(n,x,y)$ denote the polynomial at node $n$.
If node $n$ is a leaf node, then we assign the polynomial $P(n,x,y) = x$ to node $n$. 
Let node $m$ be an internal (non-leaf) node with $k$ child nodes $n_1,n_2,...,n_k$.
The polynomial at node $m$ is $P(m,x,y) = y + \Pi_{i = 1}^kP(n_i,x,y)$.
We say that the {\em topology} of a dependency tree is the tree structure without any labels.
In Figure~\ref{f2}, we show the recursive process for computing the polynomials representing the topologies of the dependency trees displayed in Figure~\ref{f1}.
It is proved that two unlabeled trees are isomorphic if and only if they have the same polynomial.
Furthermore, each term in the polynomial of an unlabeled tree is interpretable and corresponds to a specific subtree of the unlabeled tree.
See \cite{Liu2021} for more details about the tree distinguishing polynomial, and see \cite{Liu2022} for distances and methods based on the polynomial to analyze tree structures. 

\subsection{Polynomial of dependency trees}

\begin{figure}[ht]
\begin{center}
\includegraphics[scale=0.9]{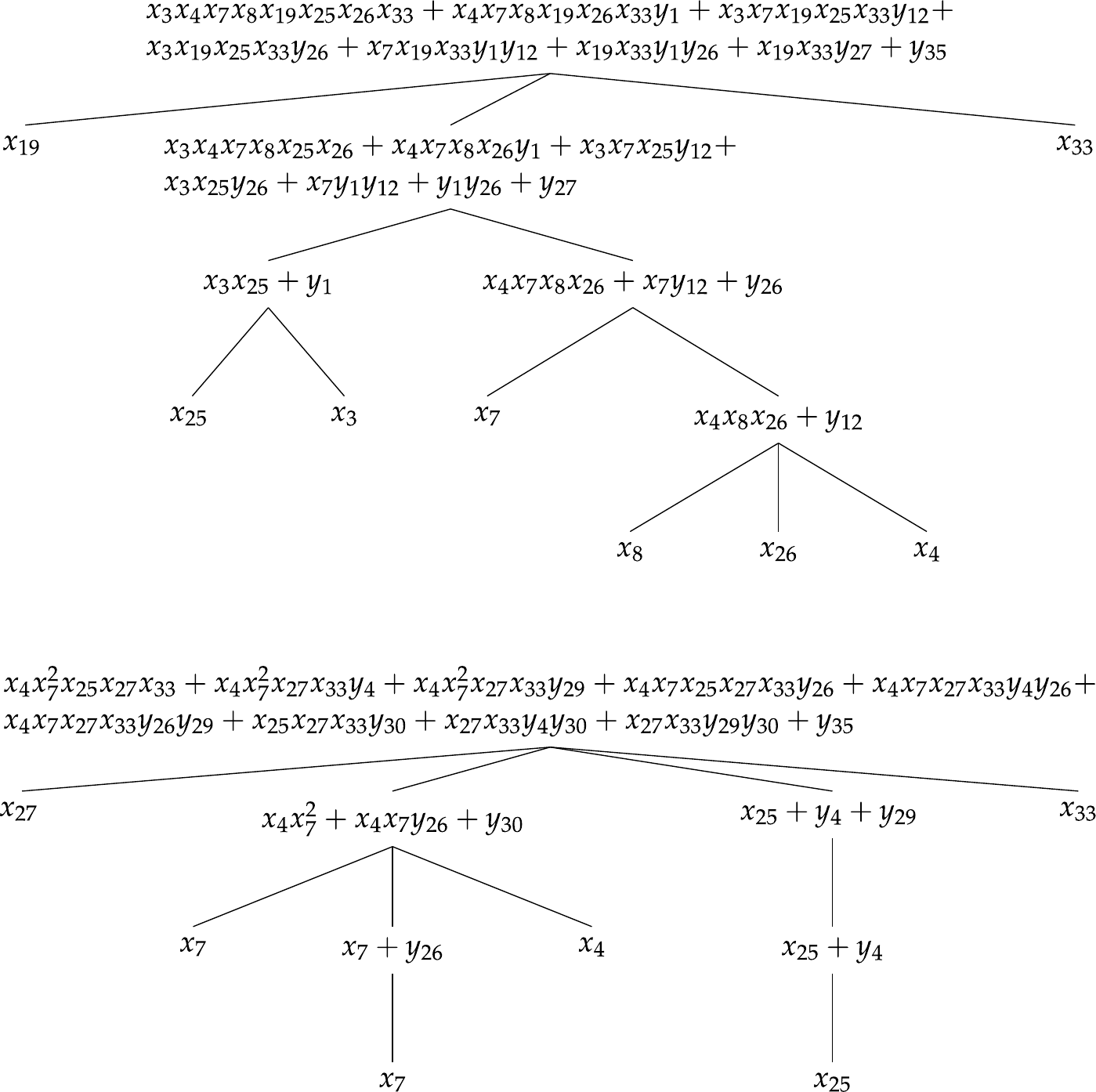}
\renewcommand{\figurename}{Figure}
\caption{{\bf Polynomials of the dependency trees.} The recursive process of computing the polynomials representing the dependency trees displayed in Figure~\ref{f1} from the leaf nodes to the roots. The polynomials at the roots represent the two dependency trees. }\label{f3}
\end{center}
\end{figure}

Here, we generalize the tree distinguishing polynomial for representing dependency trees.
Compared with tree topologies, dependency trees have node labels. 
In this study, there are 37 labels representing head-dependent relations listed in Table~\ref{t1}. These labels may appear in both leaf nodes and internal nodes of dependency trees.
So, we represent dependency trees using a generalized tree distinguishing polynomial with 74 variables classified into two sets: $X = \{x_1,x_2,...,x_{37}\}$ and $Y = \{y_1,y_2,...,y_{37}\}$.
We denote the generalized polynomial for a dependency tree $T$ by $P(T,X,Y)$.
Similarly, we compute the polynomial $P(T,X,Y)$ recursively from the leaf nodes to the root for the dependency tree $T$.
Suppose that node $n^{\ell}$ is a leaf node with label $\ell$, then we assign the polynomial $P(n^{\ell},X,Y) = x_{\ell}$ to the leaf node.
Let node $m^{\ell}$ be an internal node with label $\ell$ which has $k$ child nodes $n_1,n_2,...,n_k$, then the polynomial at node $m^{\ell}$ is $P(m^{\ell},x,y) = y_{\ell} + \Pi_{i = 1}^kP(n_i,x,y)$. 
Figure~\ref{f3} shows the process of recursively computing the generalized polynomials representing the two dependency trees displayed in Figure~\ref{f1}.
Since this is a generalization of the polynomial that distinguishes unlabeled trees, two dependency trees have the same generalized polynomial if and only if they are isomorphic and corresponding nodes have the same labels.
Therefore, Two sentences have exactly the same dependency structure if and only if the generalized polynomials of the dependency trees of the sentences are identical. 
For simplicity, we call the generalized polynomial of the dependency tree of a sentence the {\em dependency tree polynomial} of the sentence.

\subsection{Polynomial distance of dependency trees}

In the polynomial representing an unlabeled tree, the information about the hierarchical structure is encoded in the coefficient and exponents of each term; see Figure~\ref{f2}. 
In the polynomial representing a dependency tree, the syntactic information is encoded mainly in the exponents of each term due to the introduction of additional variables; see Figure~\ref{f3}.
We develop a new measure to compare dependency tree polynomials, hence the dependency trees.
The polynomial $P(T,X,Y)$ representing a dependency tree $T$ can be described term by term. 
We write each term of the polynomial as a vector with 75 entries $t = [e_{x_1},e_{x_2},...,e_{x_{37}},e_{y_1},e_{y_2},...,e_{y_{37}},c]$, where the exponent of variable $x_i$ is $e_{x_i}$, the exponent of variable $y_i$ is  $e_{y_i}$ and the coefficient of the term is  $c$. 
We call such a vector a {\em term vector} of the polynomial $P(T,X,Y)$. 
Let $P$ and $Q$ be two dependency tree polynomials and $\M_P$ and $\M_Q$ be the corresponding sets of term vectors of $P$ and $Q$.
We denote the number of term vectors in $\M_P$ (or $\M_Q$) by $\abs{\M_P}$ (or $\abs{\M_Q}$).
Let $s$ and $t$ be two term vectors. 
We denote the Manhattan distance \cite{Craw2010} between $s$ and $t$ by $\|s-t\|_1$ and define the {\em polynomial distance} for the pair of dependency tree polynomials $P$ and $Q$ using Formula (\ref{e1}).

\begin{equation}\label{e1}
    d(P,Q) = \frac{\displaystyle\sum_{s\in \M_P}\min_{t\in \M_Q}\|s-t\|_1 +\sum_{t\in \M_Q}\min_{s\in \M_P}\|s-t\|_1 }{\abs{\M_P}+\abs{\M_Q}}
\end{equation}

Since polynomials and dependency trees are in one-to-one correspondence, the defined distance for dependency tree polynomials is also for dependency trees. Without ambiguity, the polynomial distance between dependency trees refers to the distance between dependency tree polynomials throughout the paper.
Furthermore, each sentence in the PUD treebanks also has a unique dependency tree constructed under the UD framework, so, without ambiguity, the polynomial distance between sentences refers to the distance between their dependency tree polynomials.

\subsection{Experiments}

We divide the 1,000 sentences of the PUD treebanks into 5 datasets based on their original languages and name the 5 datasets using the capital ISO 639-2/B codes of the sentences' original languages.
Throughout the paper, the capital ISO 639-2/B codes of languages only refers to the 5 datasets.
The ENG dataset consists of 750 sentences originally written in English, and every sentence has 20 dependency trees corresponding to its translations in the 20 languages.
So, the ENG dataset contains 15000 dependency trees in total. 
The GER dataset consists of 100 sentences originally written in German, and every sentence has 20 dependency trees corresponding to its translations in the 20 languages.
So, the GER dataset contains 2000 dependency trees in total. 
The FRE dataset consists of 50 sentences originally written in French, and every sentence has 20 dependency trees corresponding to its translations in the 20 languages.
So, the FRE dataset contains 1000 dependency trees in total. 
The ITA dataset consists of 50 sentences originally written in Italian, and every sentence has 20 dependency trees corresponding to its translations in the 20 languages.
So, the ITA dataset contains 1000 dependency trees in total. 
The SPA dataset consists of 50 sentences originally written in Spanish, and every sentence has 20 dependency trees corresponding to its translations in the 20 languages.
So, the SPA dataset contains 1000 dependency trees in total. 
Throughout the paper, results based on different datasets are visualized in different colors. 
Results based on the ENG dataset are in blue; results based on the GER dataset are in yellow;
results based on the FRE dataset are in purple;
results based on the ITA dataset are in green; and results based on the SPA dataset are in red.

Note that every sentence in the 5 datasets is written in 20 languages, and 20 dependency trees are constructed for each sentence based on the original sentence and its 19 translations.
So, for each of the 5 datasets, a dependency tree can be identified by the original sentence and the language to which the original sentence is translated.
For each dataset, we compute the polynomials of all the dependency trees, and calculate the pairwise polynomial distances between the 20 dependency trees for every sentence.
We analyze the syntax of the sentences whose polynomial distances between a pair of translations are the smallest and the largest.
For each sentence, the pairwise distances between the 20 dependency trees form a $20\times 20$ distance matrix, which we call the {\em translation distance matrix} of the sentence. 
We say that the distance stored in each entry of the translation distance matrix of a sentence is the {\em translation distance} of the sentence between the corresponding languages of the entry.
We take the mean value of each entry in the translation distance matrices over all sentences in a dataset and call the resulting matrix the {\em language distance matrix} of the dataset.
We say that an entry in the language distance matrix of a dataset is the {\em pairwise language distance} between the corresponding pair of languages in the dataset.
The numeric value at each entry of the language distance matrix of a dataset indicates syntax similarity of a pair of languages based on the sentences in the dataset.
We summarize the language distance matrices of the 5 datasets by showing the mean and median of all pairwise language distances and pairs of the nearest and farthest languages in the pairwise language distance.
We also take the mean value of the pairwise language distances between a language and other 19 languages and call the mean value the {\em average language distance} of the language.
We show the languages with smallest and largest average language distances in the 5 datasets. 
We use the language distance matrices of the 5 datasets to perform a syntactic typology study of the 20 available languages in the PUD treebanks.
We visualize the language distance matrices using multidimensional scaling (MDS) \cite{Cox2001}, and we construct dendrograms by applying the unweighted pair group method with arithmetic mean (UPGMA) method to the language distance matrices \cite{Sokal1958}.
These visualizations provide different perspectives for analyzing syntax similarity of languages based on the sentences in the PUD treebanks.
Lastly, for each dataset, we consider the translations of all sentences in a language as a corpus of the language, and there are 20 corpora for each dataset.
We calculate all pairwise distances between translated sentences in each of the 20 corpora, and we call such a pairwise distance a {\em pairwise sentence distance} in the corpus.
We show the distribution of pairwise sentence distances for each corpus, and we call the maximum pairwise sentence distance in a corpus the {\em diameter} of the corpus.  
The diameter is a simple measure of diversity \cite{Bryant2012}, and we discuss the potential of the polynomial methods in measuring syntax diversity.

\section{Results}
\label{S:3}

\subsection{Syntax comparison of sentences}

The newly defined distance of dependency tree polynomials provides a quantitative measure of sentences' syntax similarity.
If two sentences have identical dependency structure, then the distance between the dependency tree polynomials of the sentences is zero. 
A smaller distance between a pair of sentences suggests that they are similar in syntax, and a larger distance between a pair of sentences suggests the syntax being more different. 
The distance between the dependency trees in Figure~\ref{f1} is 5.06.

\begin{figure}[!ht]
\begin{center}
\includegraphics[scale=0.52]{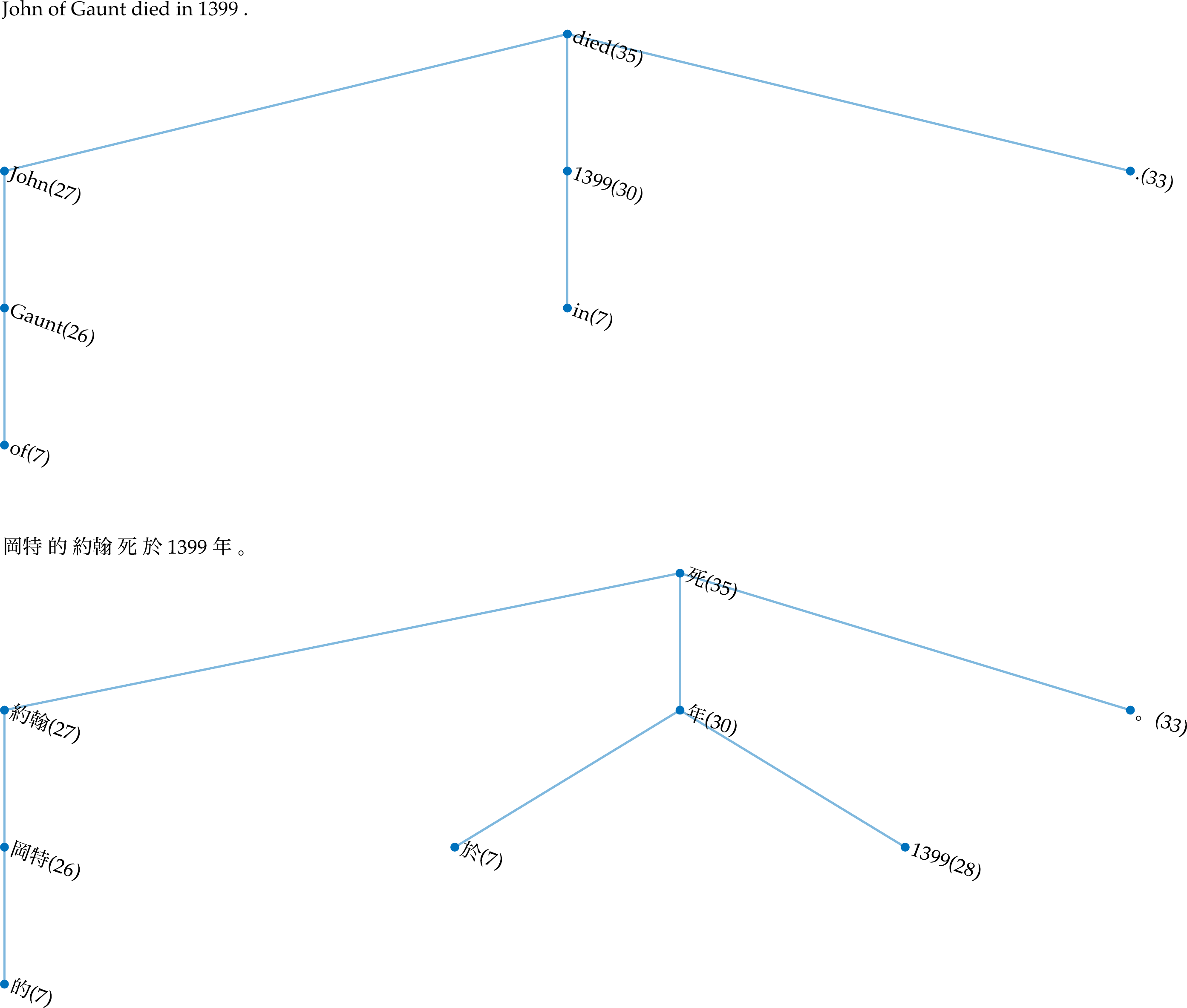}
\renewcommand{\figurename}{Figure}
\caption{{\bf The sentence in the ENG dataset with minimum polynomial distance between its English and Chinese translations.} Top: the dependency tree of the sentence's English translation (the original sentence since it is in the ENG dataset). Bottom: the dependency tree of the sentence's Chinese translation. The polynomial distance between the dependency trees is 0.43. }\label{f4}
\end{center}
\end{figure}

In Figure~\ref{f4}, we display the dependency tree of an English sentence in the ENG dataset and the dependency tree of its Chinese translation in the dataset.
The sentence's English and Chinese translations have a polynomial distance 0.43, which is the minimum distance over all sentences in the ENG dataset when comparing the distance between their English and Chinese translations.
The English sentence and the Chinese translation are syntactically similar. 
The stems of both sentences are in subject-predicate form, and the subjects of both sentences are complex noun phrases.
The only difference between the sentences is at time adverbials, where the Chinese sentence has a word after ``1399'' to indicate that the numeral represents a year.

\begin{figure}[!ht]
\begin{center}
\includegraphics[scale=0.52]{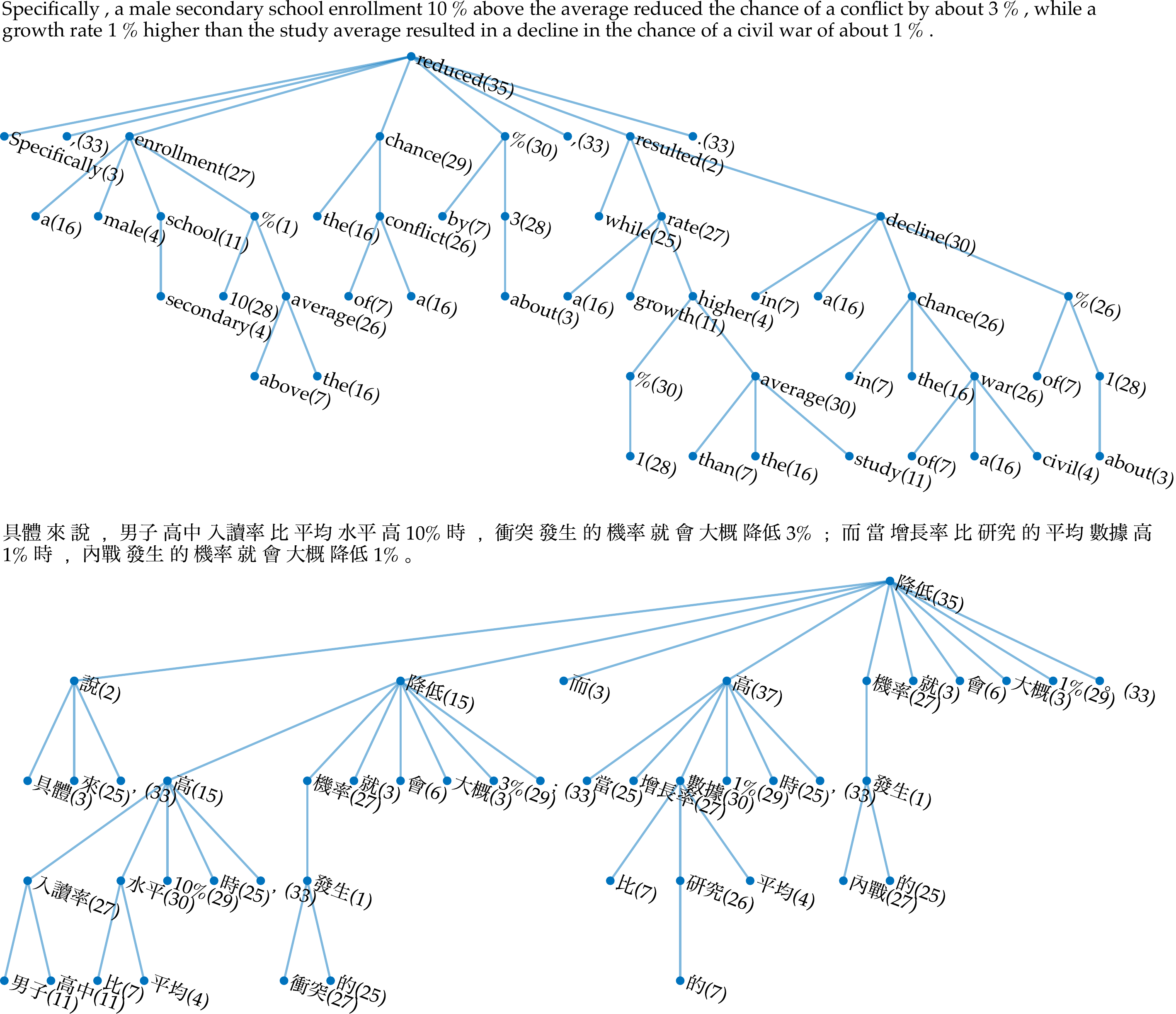}
\renewcommand{\figurename}{Figure}
\caption{{\bf The sentence in the ENG dataset with maximum polynomial distance between its English and Chinese translations.} Top: the dependency tree of the sentence's English translation (the original sentence since it is in the ENG dataset). Bottom: the dependency tree of the sentence's Chinese translation. The polynomial distance between the dependency trees is 22.93.  }\label{f5}
\end{center}
\end{figure}

In Figure~\ref{f5}, we display the dependency tree of an English sentence in the ENG dataset and the dependency tree of its Chinese translation in the dataset.
The sentence's English and Chinese translations have a polynomial distance 22.93, which is the maximum distance over all sentences in the ENG dataset when comparing the distance between their English and Chinese translations.
The English sentence and the Chinese translation have more distinct syntax from branches to the stem.
%ADD DETAIL.
The dependency tree of the English sentence is right-branching, that is, there are more modifiers to the right of the root; while the dependency tree of the Chinese translation is left-branching. 
This difference between English and Chinese is observed in other long sentences in the ENG dataset.
In terms of sentence stems, the English sentence has a double-object structure, with ``chance'' and ``'3\%'' as its objects; the Chinese translation has a single-object structure, with only ``1\%'' as its object. 
It is worth noting that the UD framework annotates percentages in Chinese and English differently. 
In Chinese, the the numeral ``1'' and the symbol ``\%'' are considered as one word which serves as the object (29) of the sentence; in English, the the numeral ``3'' and the symbol ``\%'' are treated as two separated words, comprising a numeric modifier (28) and a oblique nominal (30) of the sentence respectively. 
Furthermore, the Chinese sentence has more adverbials (3) and auxiliaries (6) directly modifying the root of the sentence.
These modifiers function in Chinese to make sentences lucid and coherent, but they are not necessary in English.
In terms of branches, the complex noun phrase ``a male secondary school enrollment 10\% above the average'' in the English sentence is expressed with a ``{\em bi}''-structure in the Chinese translation, which can be directly translated back to English as ``When a male secondary school enrollment is 10\% higher than the average''.
The  ``{\em bi}''-structure used to compare the ``enrollment'' and ``the average'' form a subject-verb-object clause, which is disparate from the complex noun phrase.
In the original English sentence, the word ``above'' is a preposition bearing a case relationship (7), while its corresponding part in the Chinese translation is the root of the clause. 
%END DETAIL.

In general, shorter sentences have fewer options for syntax variation, hence the polynomial distances between shorter sentences are more likely to be small.
In contrast, longer sentences have more room for different syntax, so the maximum polynomial distance is more likely between longer sentences.
We also display sentences in the ENG dataset with minimum and maximum polynomial distances from the original sentences to their French and Spanish translations. See Supplementary Figure~\ref{sf1}-\ref{sf4}. 

\subsection{Syntactic similarity of languages}

\begin{figure}[!ht]
\begin{center}
\includegraphics[scale=0.6825]{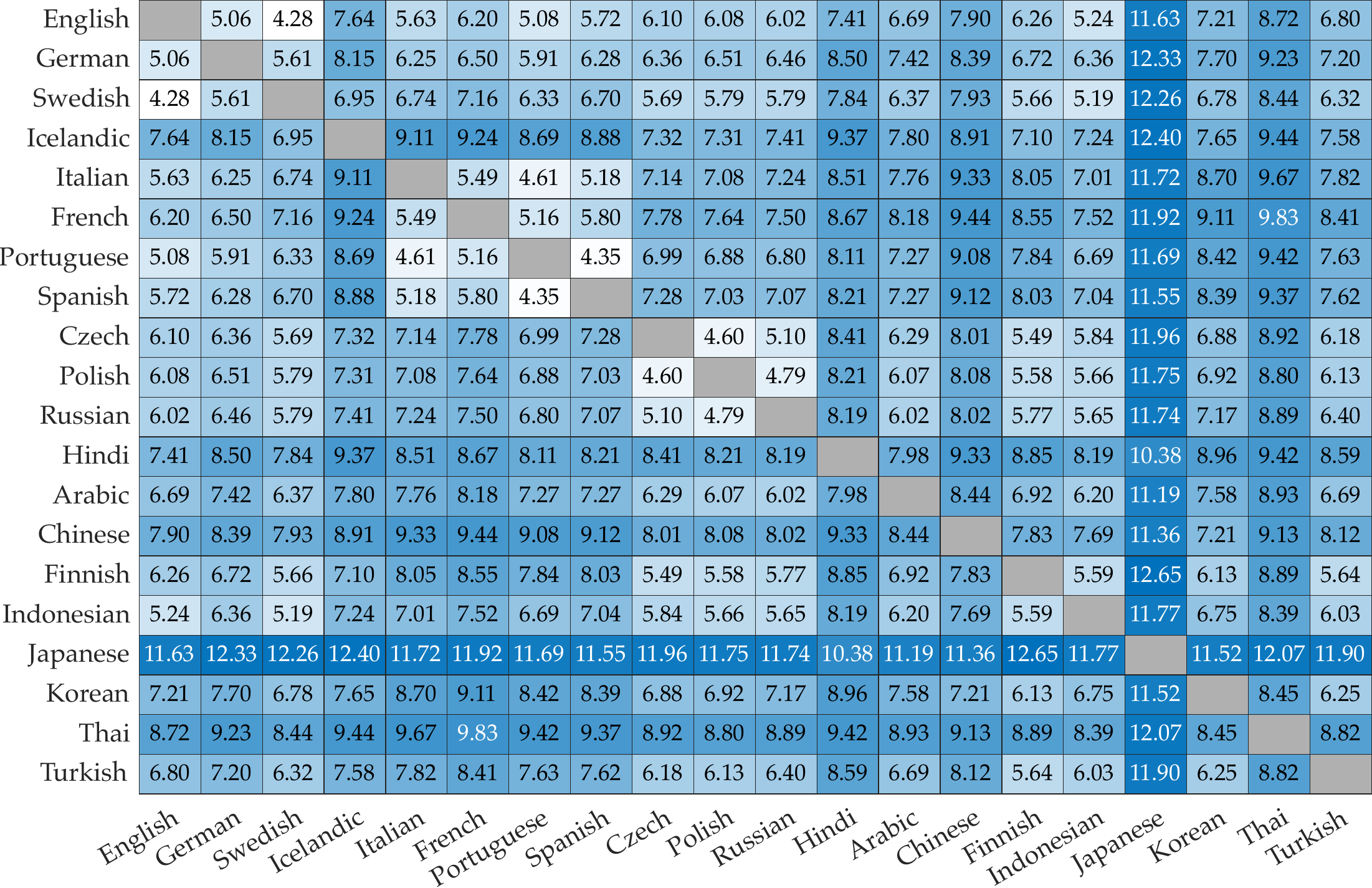}
\renewcommand{\figurename}{Figure}
\caption{{\bf The language distance matrix of the ENG dataset.} The languages are ordered based on Glottolog 4.6 classification \cite{Forkel2022}: Indo-European languages are listed first and grouped according to their subclasses (Germanic, Italic, Balto-Slavic and Indo-Iranian), and other languages are following in the alphabetical order. }\label{f6}
\end{center}
\end{figure}

\begin{figure}[!p]
\begin{center}
\includegraphics[scale=0.6825]{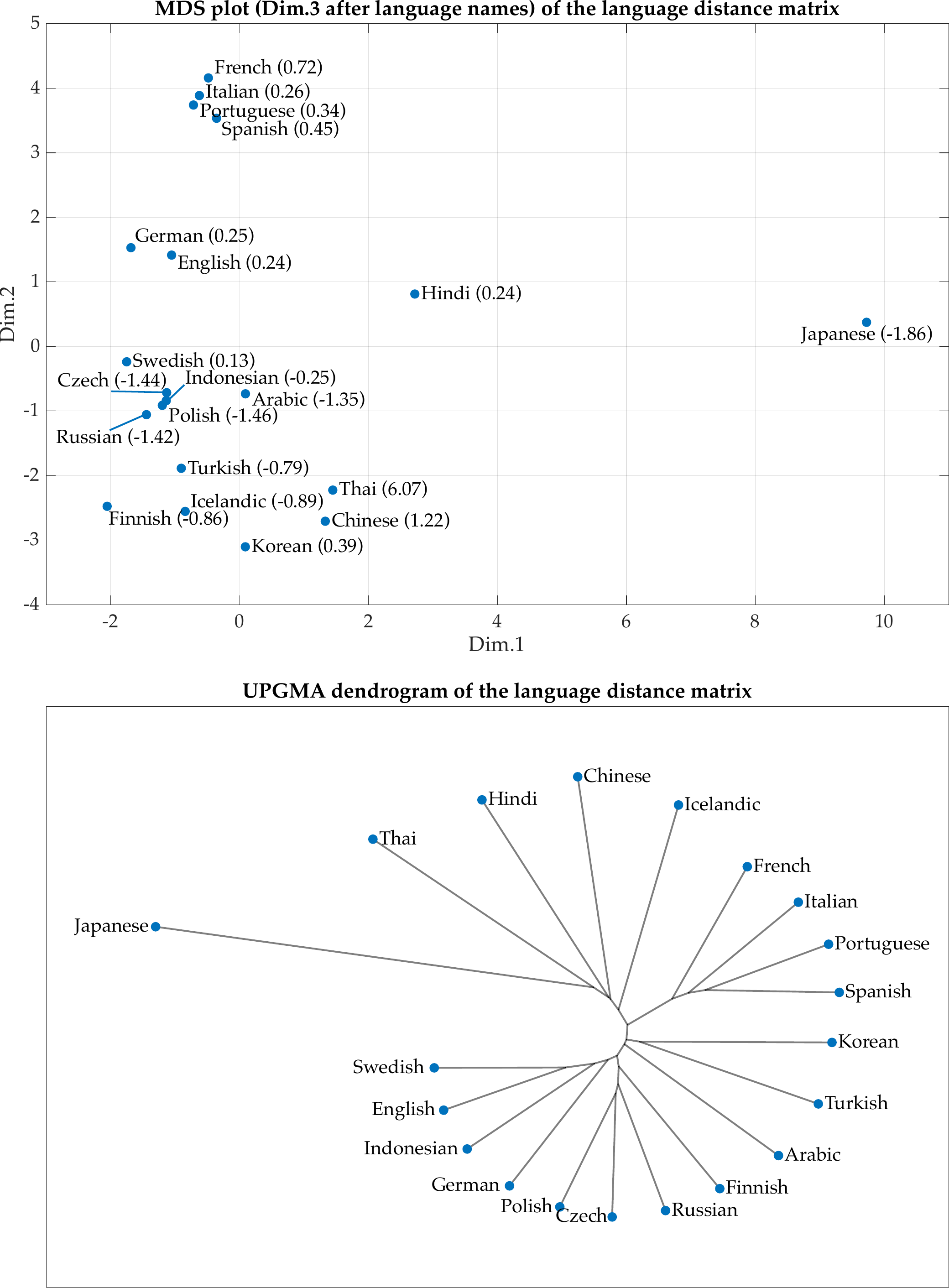}
\renewcommand{\figurename}{Figure}
\caption{{\bf Visualizations the language distance matrix of the ENG dataset.} Top: the multidimensional scaling plot of the language distance matrix of the ENG dataset. Bottom: the UPGMA dendrogram constructed based on the language distance matrix of the ENG dataset. }\label{f7}
\end{center}
\end{figure}

We show the language distance matrix of the ENG dataset in Figure~\ref{f6}, and we visualize the language distance matrix by its multidimensional scaling (MDS) plot \cite{Cox2001} and its unweighted pair group method with arithmetic mean (UPGMA) dendrogram \cite{Sokal1958}, which are displayed in Figure~\ref{f7}. 
We observe that the clustering of language similarity based on the PUD treebanks and the polynomial distance is in general consistent with the genealogical classification of languages (Glottolog 4.6) based on available historical-comparative research \cite{Forkel2022}.
In the following paragraphs, we describe syntax similarity of languages based on the language distance matrices of the 5 datasets.
All similarity and closeness are based on the current PUD treebanks and limited to the 20 available languages.

Italic languages (French, Italian, Portuguese and Spanish) are close to each other in pairwise language distance. 
This can be visualized in both the MDS plot and the UPGMA dendrogram; see Figure~\ref{f7}.
The mean pairwise language distance in the ENG dataset is 7.73; see Table~\ref{t2}.
We use mean pairwise language distances as references for syntax similarity between languages:
Languages with smaller pairwise language distances are considered similar in syntax, and languages with larger pairwise language distances are considered distinct in syntax.
The pairwise language distances between Italic languages are from 4.35 to 5.80, all smaller than the mean value 7.73.
The nearest languages to Italian are Portuguese and Spanish; the nearest languages to French are Portuguese and Italian; the nearest languages to Portuguese are Spanish and Italian; and  the nearest languages to Spanish are Portuguese and Italian.
Actually, Portuguese and Spanish are among the pairs of languages with the smallest pairwise language distance based on available PUD treebanks; see Table~\ref{t2}.
The farthest languages to Italic languages are Japanese, Thai, Chinese and Icelandic.
Based on the polynomial distance, the syntax difference between Italic languages and Icelandic is larger than the syntax difference between Chinese and English.
These are consistent in the 5 datasets; see Supplementary Figure~\ref{sf5}-\ref{sf12} and Supplementary Table~\ref{st2}.

Balto-Slavic languages (Czech, Polish and Russian) are close to each other in pairwise language distance. 
The MDS plot and the UPGMA dendrogram in Figure~\ref{f7} show that the three languages are clustered and surrounded by other languages including Arabic, Finnish, Indonesian and Swedish.
The pairwise language distances between Balto-Slavic languages are from 4.60 to 5.10, all smaller than the mean value 7.73.
The two nearest languages to each Balto-Slavic language are always the other two Balto-Slavic languages.
The nearest language to Czech is Polish and the nearest to Polish is Czech, while the nearest to Russian is Polish.
Based on available PUD treebanks, Czech and Polish are among the pairs of languages with the smallest pairwise language distance, and Balto-Slavic languages are also among the languages with the smallest average language distances, suggesting that their syntax is on average least different to all other available languages in PUD treebanks; see Table~\ref{t2}.
The farthest languages to  Balto-Slavic languages include Japanese, Thai, Chinese and Hindi.
Note that Hindi is also an Indo-European language.
We observe that the nearest language to Hindi is English with language distance 7.41, slightly smaller than the mean value 7.73 but larger than the language distance between Arabic and English, suggesting the syntax difference between English and Hindi is larger than between Arabic and English.
The farthest languages to Hindi include Japanese, Thai, Chinese and Icelandic. 
These are consistent in the 5 datasets; see Supplementary Figure~\ref{sf5}-\ref{sf12} and Supplementary Table~\ref{st3}.

For Germanic languages (English, German, Swedish and Icelandic), the pairwise language distances between English, German and Swedish are from 4.28 to 5.61, smaller than the mean value 7.73, but the pairwise language distances from Icelandic to English and German are 7.63 and 8.15, close to or larger than the mean value. 
The pairwise language distance between Icelandic and German is larger than the distance between Chinese and English in all 5 datasets; see Figure~\ref{f6} and Supplementary Figure~\ref{sf5}-\ref{sf12}.
The nearest language to German is English in all 5 datasets, and the nearest languages to Swedish are English, Indonesian and Czech; see Supplementary Table~\ref{st1}.
The nearest language to English is Swedish in the ENG dataset, which is also the smallest pairwise language distance for the ENG dataset; see Table~\ref{t2}.
However, the nearest languages to English are inconsistent in the 5 datasets, and other nearest languages include Italian, Portuguese, Spanish and German; see Supplementary Table~\ref{st1}.
According to Table~\ref{t2}, English and Swedish are among the languages with the smallest average language distances based on currently available PUD treebanks.
Icelandic also has inconsistent nearest languages in the 5 datasets, and the nearest languages include Swedish, Finnish and Balto-Slavic languages.
In the ENG dataset, the two nearest languages to Icelandic are Swedish and Finnish with distance 6.95 and 7.10, which are close to the mean value 7.73.
The farthest languages to Germanic languages include Japanese, Thai, Chinese and Hindi.
For English and German, Icelandic is among the three farthest languages, and for Icelandic, French is among the three farthest languages; see Supplementary Table~\ref{st1}.

Japanese is consistently the language with the largest average language distance in the 5 datasets based on the currently available PUD treebanks; see Table~\ref{t2}.
This is also observable in the visualizations of language distance matrices displayed in Figure~\ref{f7} and Supplementary Figure~\ref{sf9}-\ref{sf12}, suggesting that the syntax of Japanese is distinct from other 19 languages in this study. 
The three largest pairwise language distances are consistently between Japanese and Finnish, Japanese and Icelandic and Japanese and German; see Table~\ref{t2}. 
Actually, Finnish, Icelandic and German are the farthest languages to Japanese; see Supplementary Table~\ref{st4}.
Among the other 19 languages, the nearest language to Japanese is Hindi, with pairwise language distance 10.38, which is larger than all pairwise language distances between other 19 languages.
Other languages near Japanese in language distance include Arabic and Chinese, though the distances suggest rather distinct syntax between the languages.

Thai is consistently the language with the second largest average language distance in the 5 datasets based on the currently available PUD treebanks; see Table~\ref{t2}.
The nearest language to Thai is Indonesian, with a pairwise language distance 8.39 in the ENG dataset, which is larger than the mean value 7.73.
This suggests that the syntax difference between Thai and Indonesian is as large as the difference between Chinese and German.
The other languages near Thai include Swedish and Korean based on the 5 datasets, though the pairwise language distances between the languages are all larger than the mean values of the datasets.
The farthest language to Thai is Japanese, and the other far languages include French and Icelandic; see Supplementary Table~\ref{st4}.

Chinese is consistently the language with the third largest average language distance in the 5 datasets based on the currently available PUD treebanks; see Table~\ref{t2}.
The nearest language to Chinese is Korean in all 5 datasets, with a pairwise language distance 7.21 in the ENG dataset, which is close to the mean value 7.73.
This suggests that the syntax difference between Chinese and Korean is as large as between English and Korean.
The other language near Chinese is Indonesian, with a pairwise language distance 7.69 in the ENG dataset, which is slightly smaller than the pairwise language distance 7.90 between Chinese and English.
The farthest language to Chinese is also Japanese, and the second farthest language is French; see Supplementary Table~\ref{st4}.

The nearest languages to Indonesian in the 5 datasets of the PUD treebanks are Swedish and English, respectively with pairwise language distances 5.19 and 5.24 in the ENG dataset.
This suggests that the syntax difference between Indonesian and Swedish or English is smaller than the difference between English and Czech; see Supplementary Table~\ref{st4}, Figure~\ref{f6} and Supplementary Figure~\ref{sf5}-\ref{sf8}.
The farthest languages to Indonesian are Japanese, Thai and Hindi.

The nearest languages to Arabic in the 5 datasets of the PUD treebanks are Balto-Slavic languages, especially Russian and Polish.
The pairwise language distance between Arabic and Russian is 6.02 in the ENG dataset, and the distance between Arabic and Polish is 6.07.
This suggests that the syntax difference between Arabic and Russian or Polish is as large as the difference between English and Russian; see Supplementary Table~\ref{st4}, Figure~\ref{f6} and Supplementary Figure~\ref{sf5}-\ref{sf8}.
The farthest languages to Arabic are Japanese, Thai and Chinese.

In Supplementary Table~\ref{st4}, we observe that the nearest language to Finnish is Czech in the PUD treebanks. Other languages near Finnish include other Balto-Slavic languages, Indonesian and Turkish. 
The pairwise language distance between Finnish and Czech is 5.49 in the ENG dataset, which is as small as the distance between Italian and French; see Figure~\ref{f6} and Supplementary Figure~\ref{sf5}-\ref{sf8}.
The farthest languages to Finnish are Japanese, Thai and Hindi. 
For Turkish, the nearest language is Finnish in all 5 datasets, and other languages near Turkish are Balto-Slavic languages or Indonesian. 
The pairwise language distance between Turkish and Finnish is 5.64 in the ENG dataset, which is as small as the distance between Italian and German; see Figure~\ref{f6} and Supplementary Figure~\ref{sf5}-\ref{sf8}.
The farthest languages to Turkish are Japanese, Thai, Hindi and French. 
Lastly, we observe that in all 5 datasets, the nearest language to Korean is Finnish and the second nearest language to Korean is Turkish, with pairwise language distance 6.13 and 6.25 in the ENG dataset respectively; see Supplementary Table~\ref{st4}.
Both distances are smaller than the mean value 7.73.
Based on the ENG, GER and FRE datasets, the syntax difference between Korean and Finnish or Turkish is as small as the syntax difference between English and French, while in the ITA and SPA datasets, the syntax difference between Korean and Finnish or Turkish is as small as the syntax difference between English and Polish; see Figure~\ref{f6} and Supplementary Figure~\ref{sf5}-\ref{sf8}.
It is also observed in the UPGMA dendrograms that the Korean, Finnish and Turkish are closely related, especially that Korean and Turkish share common ancestry in the dendrograms of ENG and SPA datasets; see Figure~\ref{f7} and Supplementary Figure~\ref{sf9}-\ref{sf12}.
This coincides with a recent unified study leveraging genetics, archaeology and linguistics to show that Korean and Turkish share common ancestry \cite{Robbeets2021}. 
However, the connection between Korean and Finnish is unclear with only initial studies discussing the similarity between the two languages \cite{Hadland1989} and studies of ancient genomics revealing the spread of Siberian ancestry in northern Europe \cite{Lamnidis2018}.

\begin{table}[!ht]
    \centering
    \small
\begin{tabular}{ |p{2.2cm}||p{2.1cm}|p{2.1cm}|p{2.1cm}|p{2.1cm}|p{2.1cm}| }
\hline
Dataset & ENG & GER & FRE & ITA & SPA  \\
 \hline
 \multicolumn{6}{|c|}{Pairwise language distance} \\
 \hline
Mean & 7.73 & 6.55 & 7.40 & 7.71 & 7.96  \\

 Median & 7.55 & 6.56 & 7.38 & 7.39 & 7.82  \\

  Smallest & 4.28  & 3.24 & 3.60 & 3.99 & 4.13  \\
 &  (eng vs swe) & (por vs spa) & (por vs spa) & (por vs spa) & (ita vs por)  \\

   2nd smallest & 4.35 & 3.44 & 4.14 & 4.21 & 4.16  \\
    &(por vs spa) & (ita vs spa) & (pol vs rus) & (cze vs pol) & (por vs spa)  \\
 
    3rd smallest & 4.60  & 3.46 & 4.39 & 4.39 & 4.34  \\
     & (czh vs pol)  & (cze vs pol) & (ita vs por) & (pol vs rus) & (fre vs ita)  \\

     3rd largest & 12.33  & 9.80 & 11.63 & 12.06 & 12.51  \\
      &  (ger vs jpn) & (ger vs jpn) & (ger vs jpn) & (ger vs jpn) & (ger vs jpn) \\

     2nd largest & 12.40  & 10.08 & 11.84 & 12.43 & 12.78  \\
  &  (ice vs jpn) & (fin vs jpn) & (ice vs jpn) & (ice vs jpn) & (ice vs jpn)  \\

     Largest & 12.65 & 10.18 & 12.22 & 12.53 & 12.96  \\
    & (fin vs jpn) & (ice vs jpn) & (fin vs jpn) & (fin vs jpn) & (fin vs jpn)  \\
 \hline
 \multicolumn{6}{|c|}{Average language distance} \\
 \hline
     Smallest & 6.61 (eng) & 5.61 (cze) & 6.40 (rus) & 6.77 (eng) & 6.85 (eng) \\

     2nd smallest & 6.73 (swe) & 5.70 (pol) & 6.47 (eng) & 6.78 (por) & 6.97 (cze)  \\

     3rd smallest & 6.85 (ind) & 5.73 (swe) & 6.47 (swe) & 6.80 (rus) & 7.05 (swe) \\

     3rd largest & 8.60 (chi) & 7.59 (chi) & 8.38 (chi) & 8.74 (chi) & 9.12 (chi)  \\

     2rd largest & 9.20 (tha) & 7.93 (tha) & 9.07 (tha) & 9.49 (tha) & 9.67 (tha)  \\

      Largest & 11.78 (jpn) & 9.53 (jpn) & 11.00 (jpn) & 11.52 (jpn) & 11.96 (jpn)  \\
 \hline
\end{tabular}
    \caption{{\bf Summaries of language distance matrices.} Language are represented by their ISO 639-2/B codes.}
    \label{t2}
\end{table}

\subsection{Syntax diversity of corpora}

\begin{figure}[ht]
\begin{center}
\includegraphics[scale=0.6825]{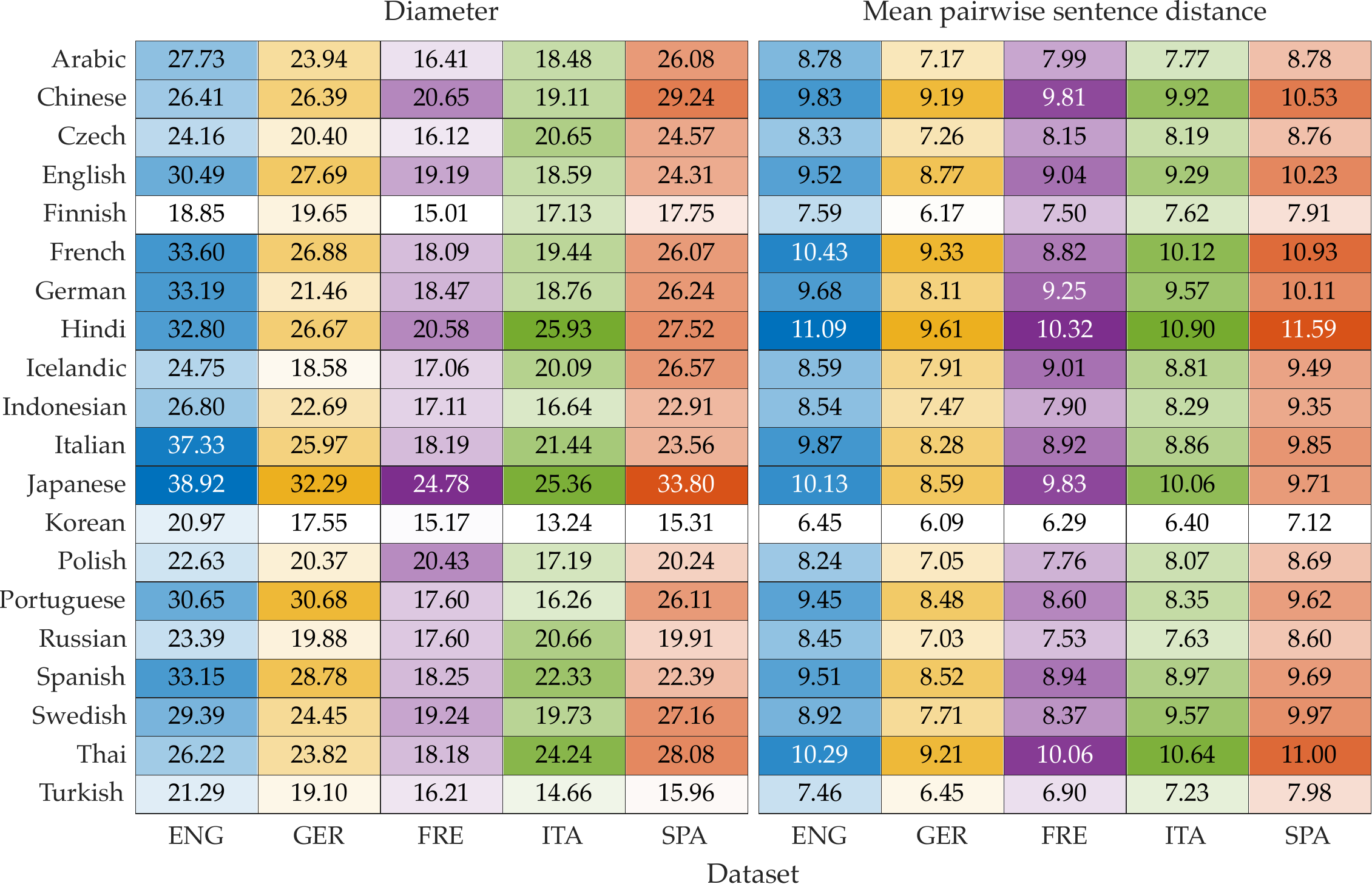}
\renewcommand{\figurename}{Figure}
\caption{{\bf The diameters and the mean pairwise sentence distances of the 20 corpora in the 5 datasets.} Every row records the diameters and mean pairwise sentence distances of the 5 corpora in the corresponding language of the 5 datasets.  }\label{f8}
\end{center}
\end{figure}

We consider the translations of all sentences in a language a corpus in a dataset.
By comparing the pairwise sentence distances of a corpus, we can describe its syntax diversity.
Here, we use two simple measures, the diameter and the mean pairwise sentence distance, to describe the syntax diversity of each dataset's 20 corpora.
Each corpus contains the translations of all sentences in the dataset, so the 20 corpora in a dataset express the same content in different languages.
The diameters and the mean pairwise sentence distances of the 5 datasets are displayed in Figure~\ref{f8}, and the detailed distributions of the pairwise sentence distances for the corpora of the 5 datasets are displayed in Supplementary Figure~\ref{sf13}-\ref{sf17}.
It is observed that the diameters and the mean pairwise distances for Finnish, Korean and Turkish are consistently smaller than other languages, and the diameters and the mean pairwise distances for Japanese and Hindi are in general larger than other languages.
This suggests that to express the same information of the corpora, Finnish, Korean and Turkish use more similar syntax, and Hindi and Japanese use more dissimilar syntax, compared with other languages. 
% This is in part because the sentences in Finnish, Korean and Turkish are short, and the sentences in Japanese are long.

\section{Discussion}
\label{S:4}

We have generalized the tree distinguishing polynomial for representing dependency trees and defined a distance between the dependency polynomials for comparing syntax of sentences. 
Compared to other methods for analyzing dependency grammar such as studying order of words \cite{Chen2017,Gerdes2021} and calculating dependency distance \cite{Chen2022,Lei2020}, the polynomial-based methods analyze dependencies from a more comprehensive perspective, taking into account all structural information and dependency relations.

The polynomial-based methods have been applied to analyze 1,000 sentences in the Parallel Universal Dependency (PUD) treebanks, and each treebank contains the translations of the 1,000 sentences in a language.
To analyze their syntax, we divided the sentences into 5 datasets based on their original languages.
We have compared the sentences with the minimum and maximum polynomial distances between their English and Chinese, French or Spanish translations.
This demonstrates the capability of comparing syntax with polynomial-based methods.
With the PUD treebanks, we have computed the average pairwise polynomial distance over all sentences in a dataset for each pair of languages.
We have used the pairwise language distance to perform a syntactic typology study of the 20 available languages, and we have conducted the analysis for all 5 datasets.
The typological results based on the 5 datasets in general agree the genealogical classification in Glottolog 4.6 \cite{Forkel2022}, though there are only 50 to 100 sentences originally written in German, French, Italian and Spanish which form the GER, FRE, ITA and SPA datasets respectively. 
With the polynomial-based methods, we have also observed less discussed syntactic typology results, for example, the connection between Finnish and Korean and a recently discussed Korean-Turkish link from a study using genetics, archaeology and linguistics \cite{Robbeets2021}. 

We have demonstrated using the polynomial distance to measure syntax diversity of corpora by showing the distributions of pairwise polynomial distances between all pairs of sentences in the corpora. 
The diameters and the mean pairwise sentence distances provide simple measures of syntax diversity of the corpora.
With proper datasets, the polynomial-based methods can be applied to, for example, measure language acquisition, assess fidelity of artificial intelligence generated text, guide artificial intelligence for generating syntactic diverse content, analyze writing styles and detect languages' syntax change over time.

With  more sentences being annotated with the Universal Dependencies framework and more Parallel Universal Dependencies treebanks being constructed, we expect that this method can reveal more information about languages, corpora and their connections and motivate new investigations in linguistics.

\section*{Implementation}
Code and data for analyses conducted in this paper are available at the repository
\url{https://github.com/pliumath/dependencies}.

\section*{Acknowledgments}
P.L. was partially supported by the grant of the Federal Government of Canada's Canada 150 Research Chair program to Prof.~C. Colijn and by the National Science Foundation DMS/NIGMS award \#2054347 to Prof. M. V\'azquez. 
R.L. was supported by Start-up funds for scientific research of BNUZ.

\bibliographystyle{abbrvnat}
\bibliography{bibliography.bib}

\clearpage
\section*{Supplementary material}

\setcounter{figure}{0}
\setcounter{table}{0}

We display the sentence in the ENG dataset with minimum and maximum polynomial distances between its English and French translations in Supplementary Figure~\ref{sf1} and \ref{sf2}.
There are more than one sentences in the ENG dataset whose English and French translations have the same dependency tree, so we only display one sentence with the minimum polynomial distance between its English and French translations.
We display the sentence in the ENG dataset with minimum and maximum polynomial distances between its English and Spanish translations in Supplementary Figure~\ref{sf3} and \ref{sf4}.
There are more than one sentences in the ENG dataset whose English and Spanish translations have the same dependency tree, so we only display one sentence with the minimum polynomial distance between its English and Spanish translations.

We display the language distance matrices of the GER, FRE, ITA and SPA datasets in Supplementary Figure~\ref{sf5}-\ref{sf8}. 
We summarize the pairwise language distances in the 5 datasets by listing the three nearest languages and the three farthest languages to each of the 20 languages in the 5 datasets. 
The summary for Germanic languages is listed in Supplementary Table~\ref{st1}; the summary for Italic languages is listed in Supplementary Table~\ref{st2}; the summary for Balto-Slavic and Hindi is listed in Supplementary Table~\ref{st3}; the summary for non-Indo-European languages is listed in Supplementary Table~\ref{st4}. 
We display the visualizations of the language distance matrices of the GER, FRE, ITA and SPA datasets in Supplementary Figure~\ref{sf9}-\ref{sf12}. 

The ENG dataset has 750 sentences, so there are 280,875 pairwise sentence distances between the translations of all 750 sentences in a language. 
We consider the 750 translations in a language as a corpus of the languages.
We display the distributions of the 280,875 pairwise sentence distances and the diameters of the 20 corpora in Supplementary Figure~\ref{sf13}.
The GER dataset has 100 sentences, so there are 4,950 pairwise sentence distances in each corpus. 
The distributions of the 4,950 pairwise sentence distances and the diameters of the 20 corpora are displayed in Supplementary Figure~\ref{sf14}.
Each of the FRE, ITA and SPA datasets has 50 sentences, so there are 1,225 pairwise sentence distances in each corpus of each dataset.
We display the distributions and diameters of the pairwise sentence distances of the 3 datasets in Supplementary Figure~\ref{sf14}-\ref{sf17} respectively.

\begin{figure}[!p]
\begin{center}
\includegraphics[scale=0.52]{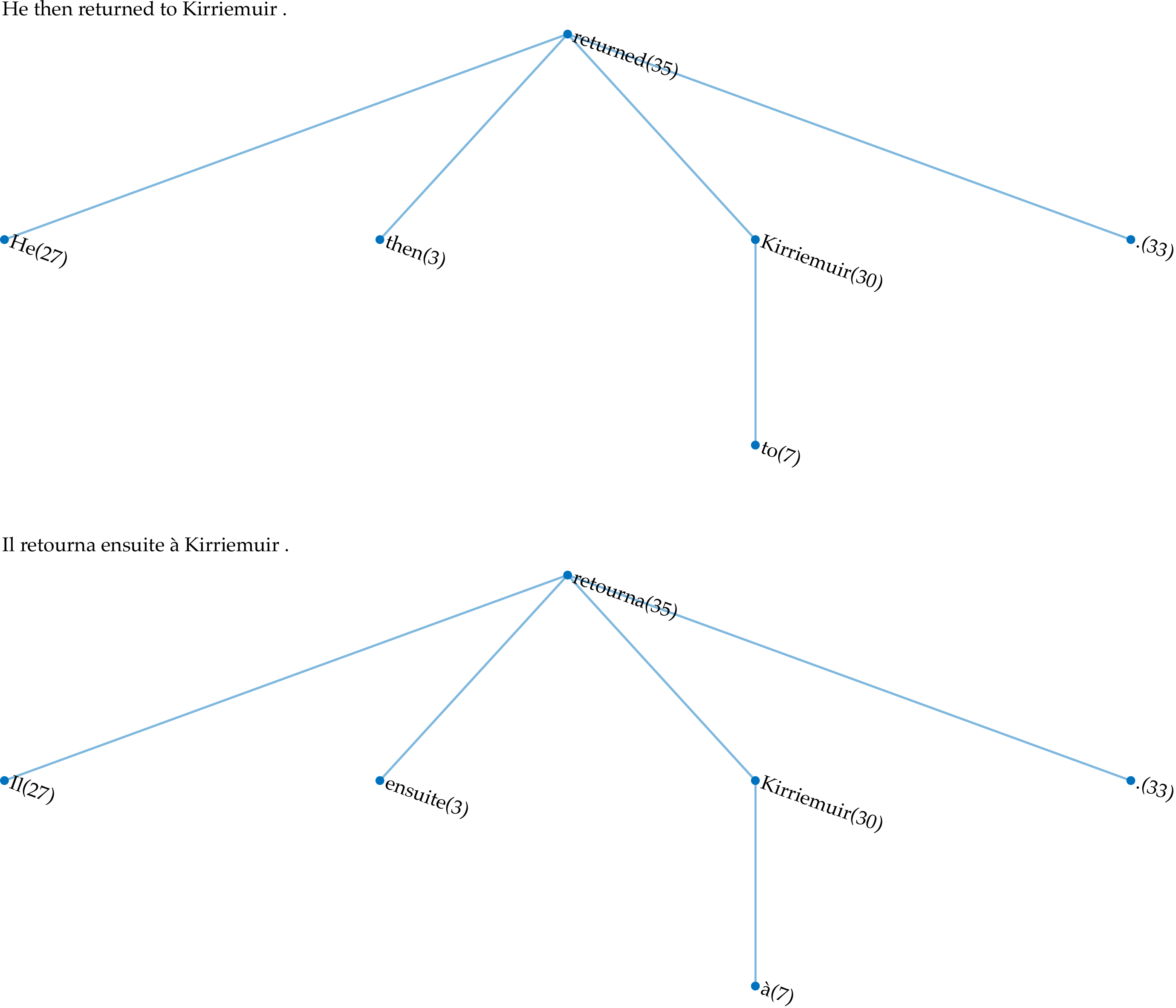}
\renewcommand{\figurename}{Supplementary Figure}
\caption{{\bf A sentence in the ENG dataset with minimum polynomial distance between its English and French translations.} Top: the dependency tree of the sentence's English translation (the original sentence since it is in the ENG dataset). Bottom: the dependency tree of the sentence's French translation. The polynomial distance between the dependency trees is 0.  }\label{sf1}
\end{center}
\end{figure}

\begin{figure}[!p]
\begin{center}
\includegraphics[scale=0.52]{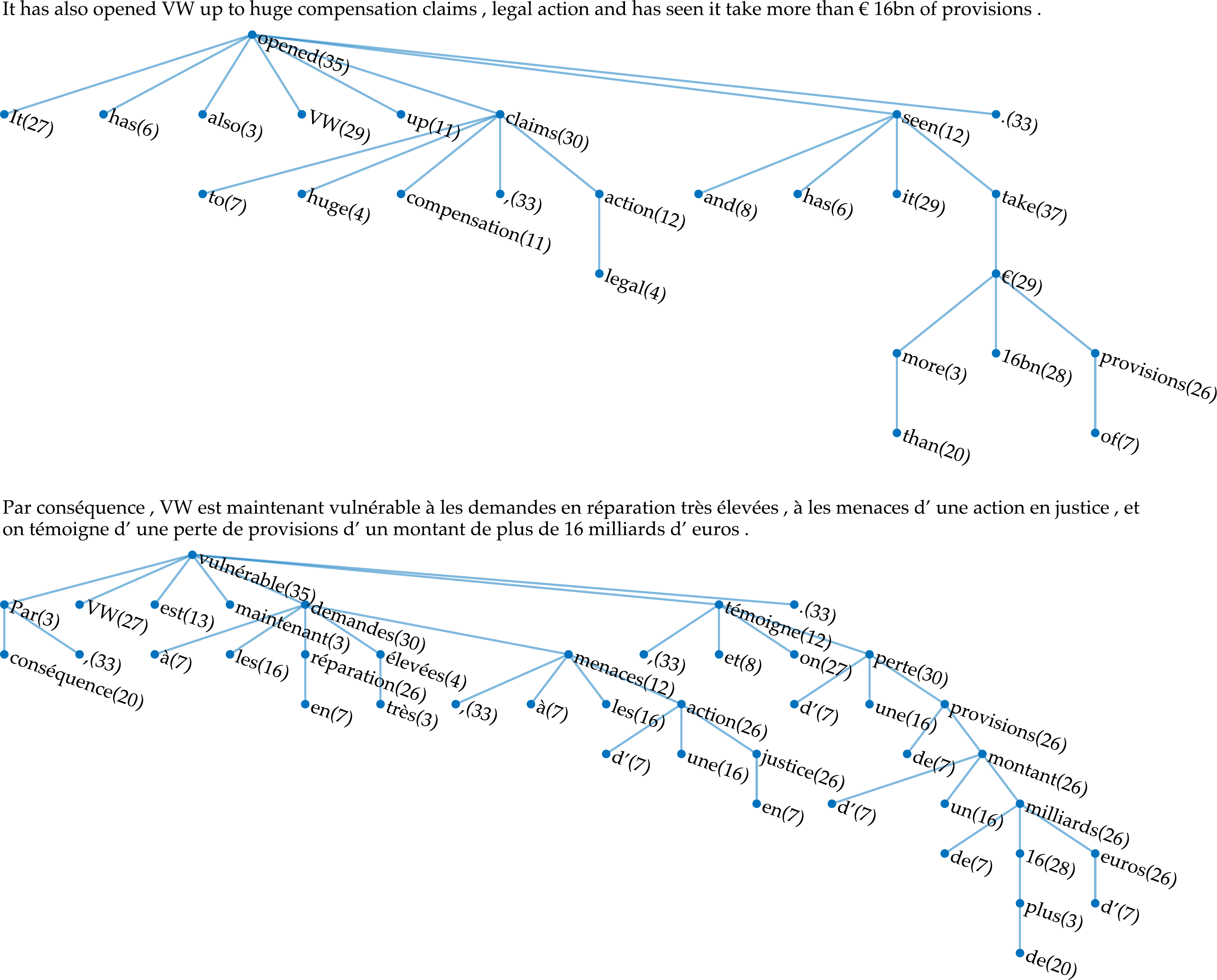}
\renewcommand{\figurename}{Supplementary Figure}
\caption{{\bf The sentence in the ENG dataset with maximum polynomial distance between its English and French translations.} Top: the dependency tree of the sentence's English translation (the original sentence since it is in the ENG dataset). Bottom: the dependency tree of the sentence's French translation. The polynomial distance between the dependency trees is 20.48.  }\label{sf2}
\end{center}
\end{figure}

\begin{figure}[!p]
\begin{center}
\includegraphics[scale=0.52]{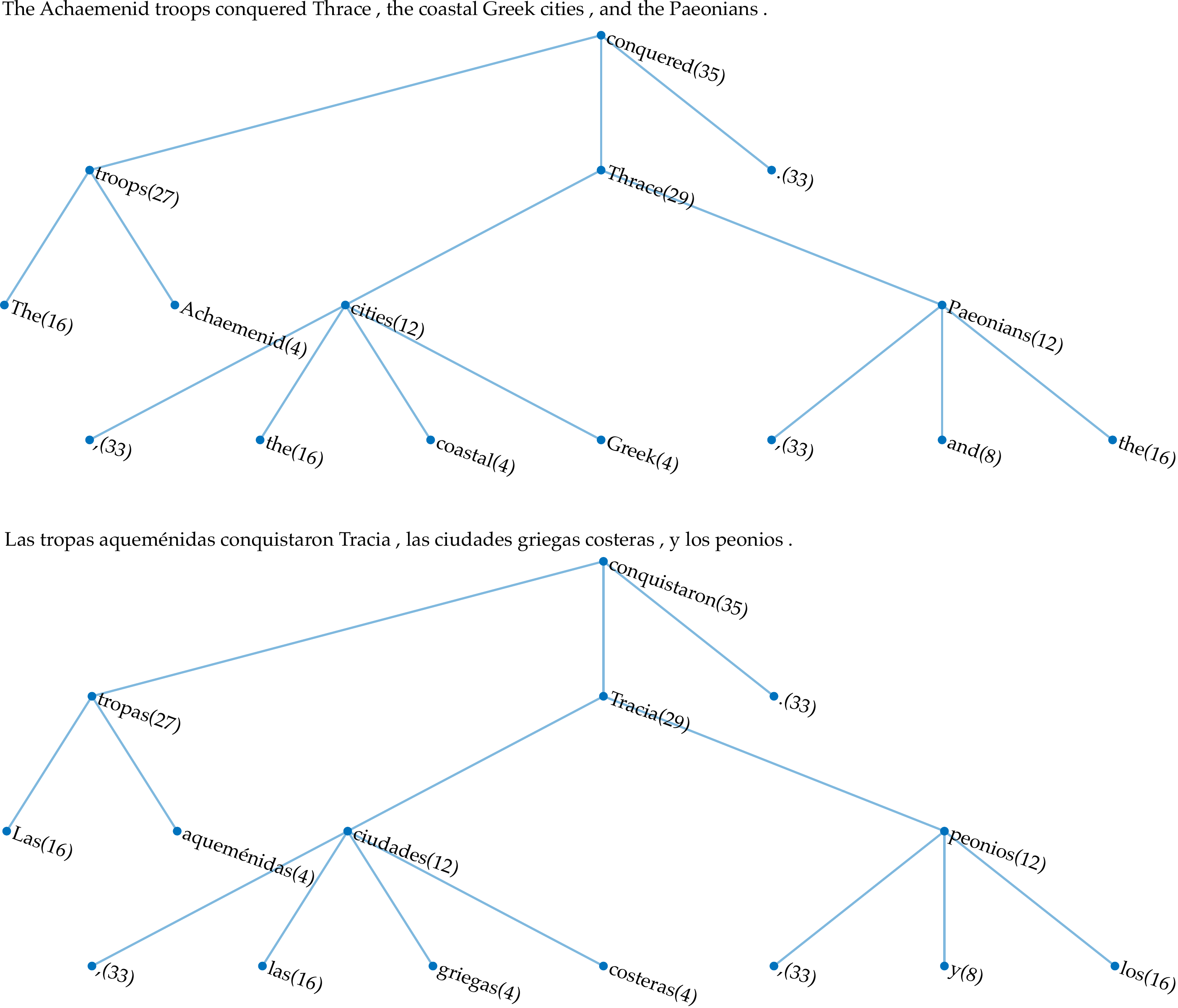}
\renewcommand{\figurename}{Supplementary Figure}
\caption{{\bf A sentence in the ENG dataset with minimum polynomial distance between its English and Spanish translations.} Top: the dependency tree of the sentence's English translation (the original sentence since it is in the ENG dataset). Bottom: the dependency tree of the sentence's Spanish translation. The polynomial distance between the dependency trees is 0.  }\label{sf3}
\end{center}
\end{figure}

\begin{figure}[!ht]
\begin{center}
\includegraphics[scale=0.52]{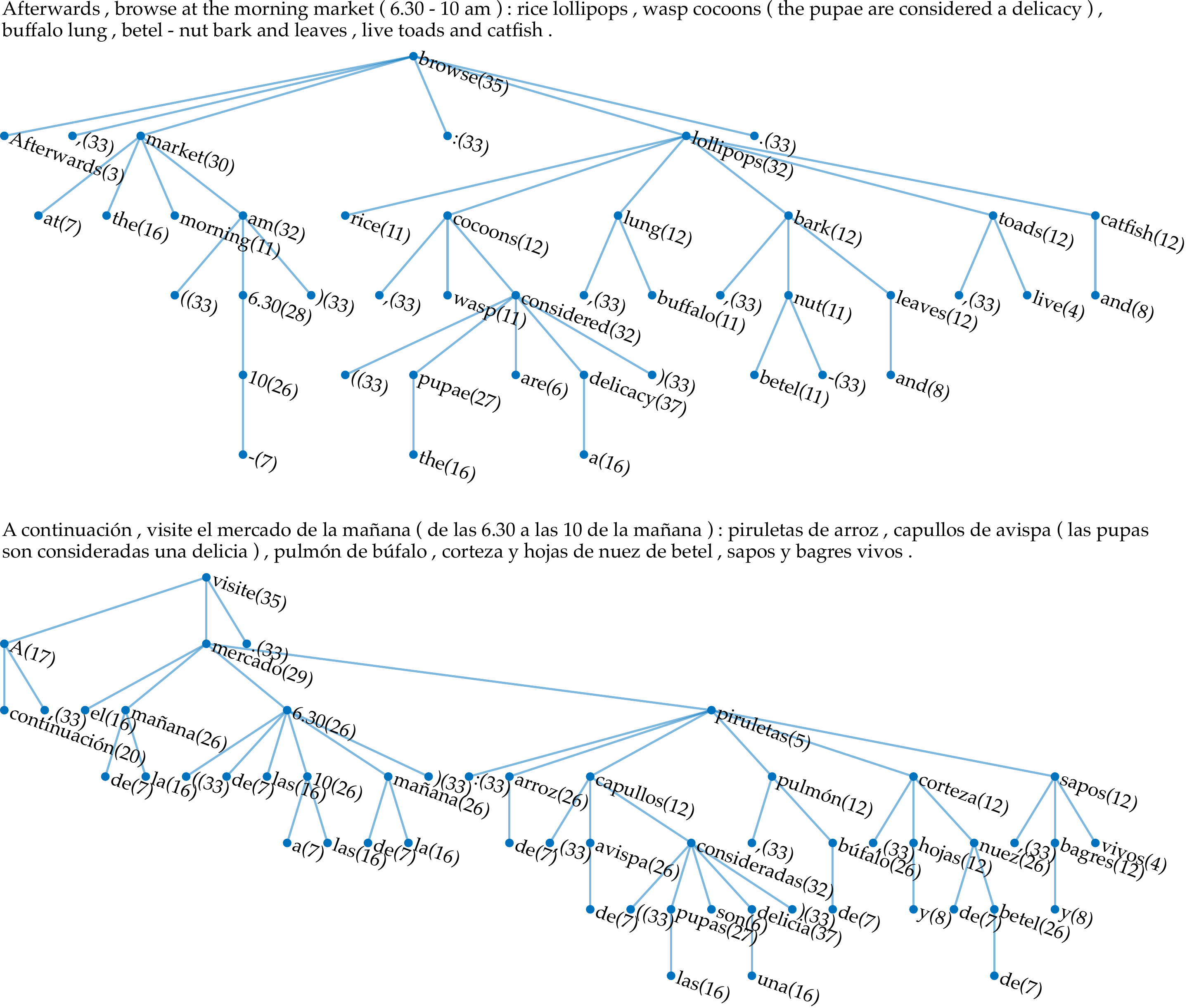}
\renewcommand{\figurename}{Supplementary Figure}
\caption{{\bf The sentence in the ENG dataset with maximum polynomial distance between its English and Spanish translations.} Top: the dependency tree of the sentence's English translation (the original sentence since it is in the ENG dataset). Bottom: the dependency tree of the sentence's Spanish translation. The polynomial distance between the dependency trees is 15.81.  }\label{sf4}
\end{center}
\end{figure}

\begin{figure}[ht]
\begin{center}
\includegraphics[scale=0.6825]{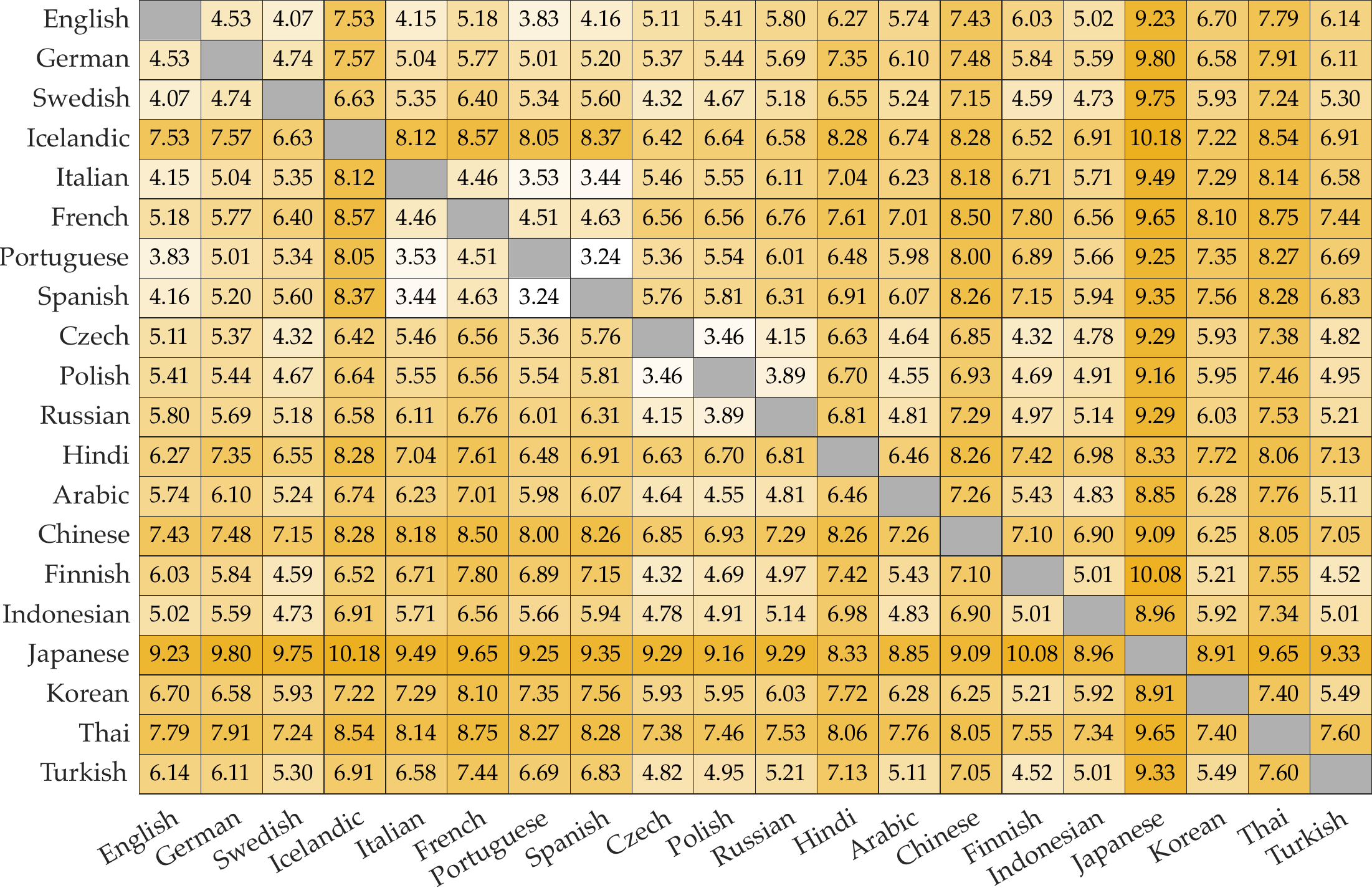}
\renewcommand{\figurename}{Supplementary Figure}
\caption{{\bf The language distance matrix of the GER dataset.} The languages are ordered based on Glottolog 4.6 classification: Indo-European languages are listed first and grouped according to their subclasses (Germanic, Italic, Balto-Slavic and Indo-Iranian), and other languages are following in the alphabetical order.  }\label{sf5}
\end{center}
\end{figure}

\begin{figure}[ht]
\begin{center}
\includegraphics[scale=0.6825]{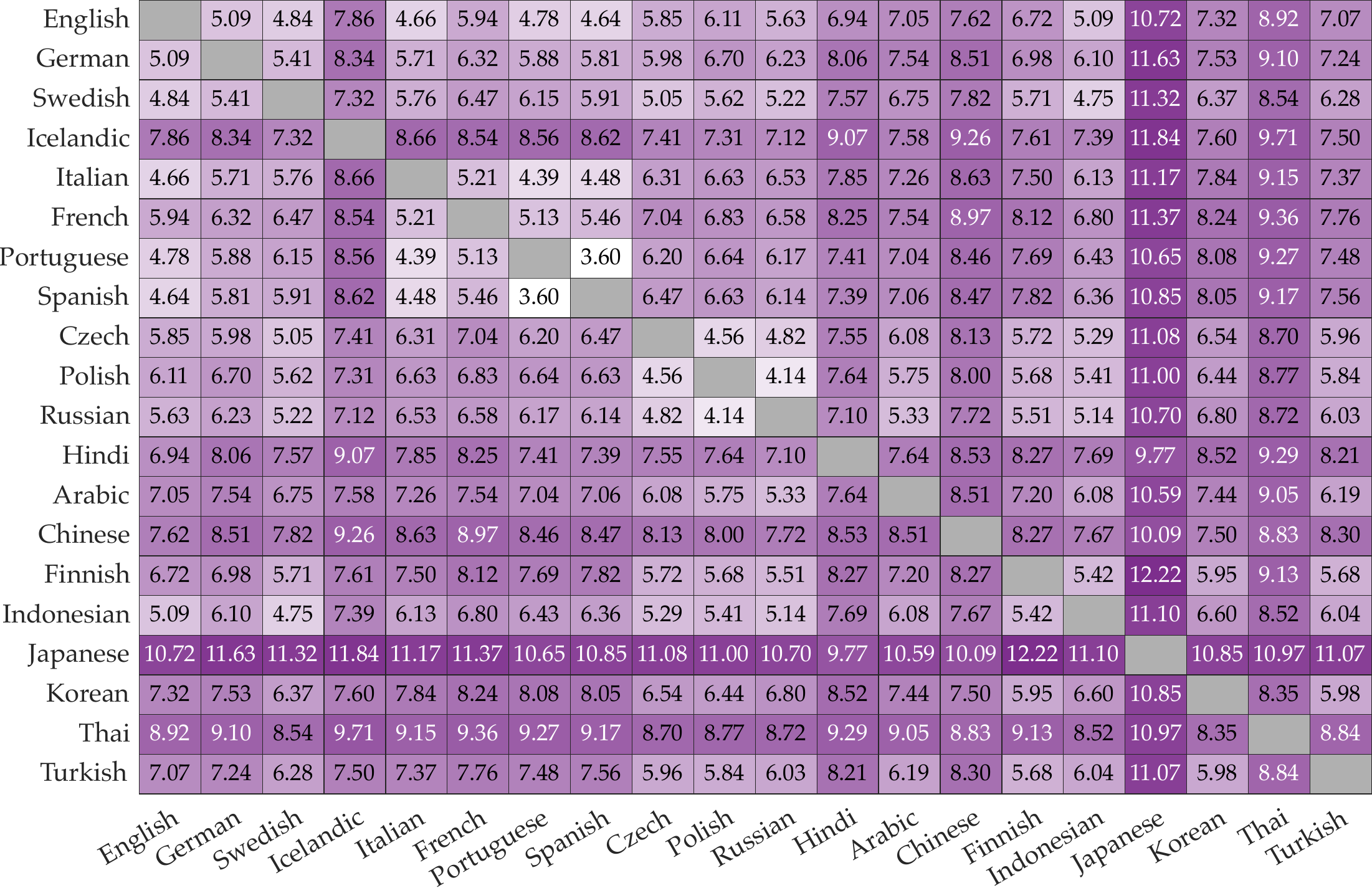}
\renewcommand{\figurename}{Supplementary Figure}
\caption{{\bf The language distance matrix of the FRE dataset.} The languages are ordered based on Glottolog 4.6 classification: Indo-European languages are listed first and grouped according to their subclasses (Germanic, Italic, Balto-Slavic and Indo-Iranian), and other languages are following in the alphabetical order.   }\label{sf6}
\end{center}
\end{figure}

\begin{figure}[ht]
\begin{center}
\includegraphics[scale=0.6825]{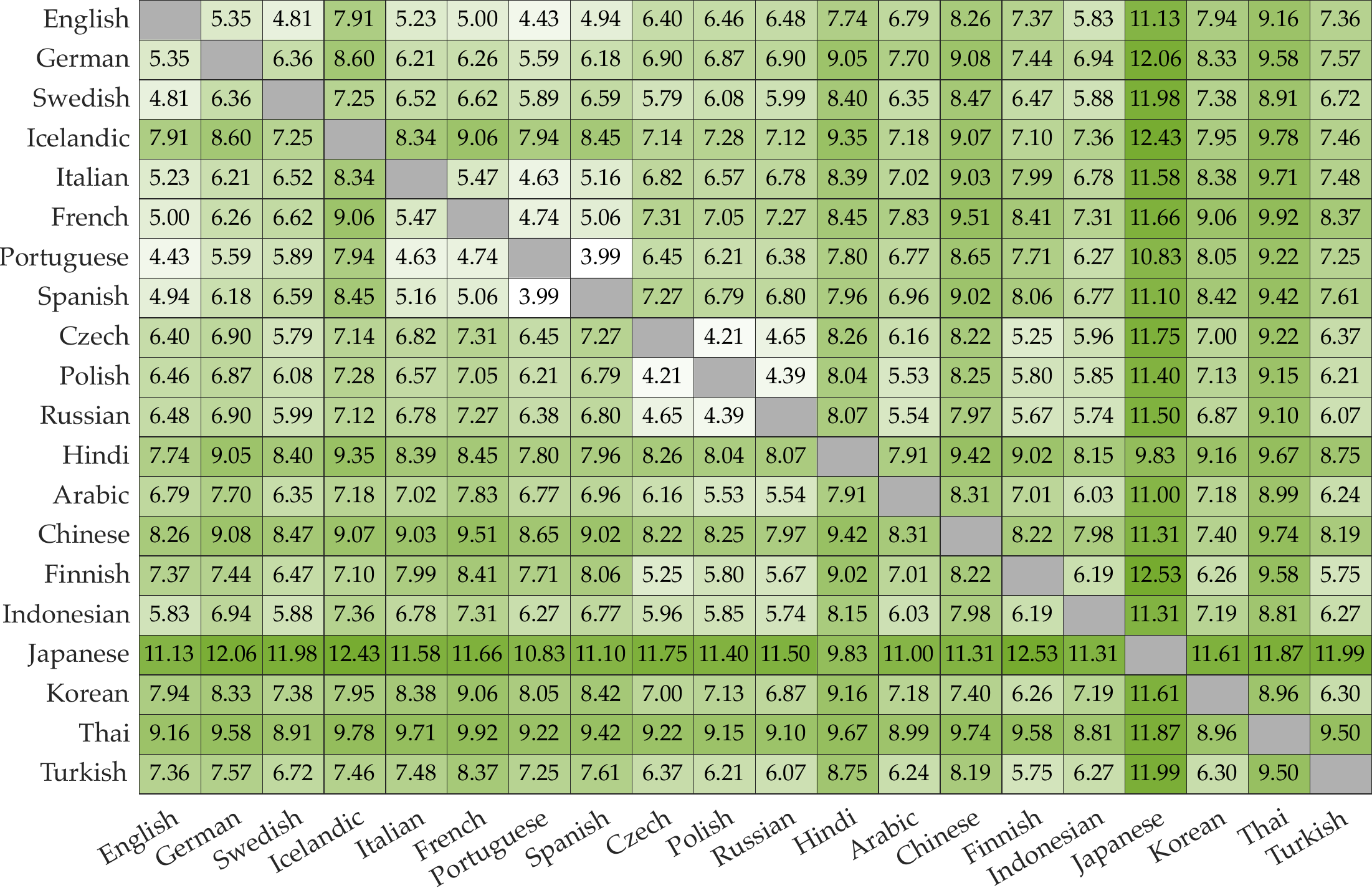}
\renewcommand{\figurename}{Supplementary Figure}
\caption{{\bf The language distance matrix of the ITA dataset.} The languages are ordered based on Glottolog 4.6 classification: Indo-European languages are listed first and grouped according to their subclasses (Germanic, Italic, Balto-Slavic and Indo-Iranian), and other languages are following in the alphabetical order.  }\label{sf7}
\end{center}
\end{figure}

\begin{figure}[ht]
\begin{center}
\includegraphics[scale=0.6825]{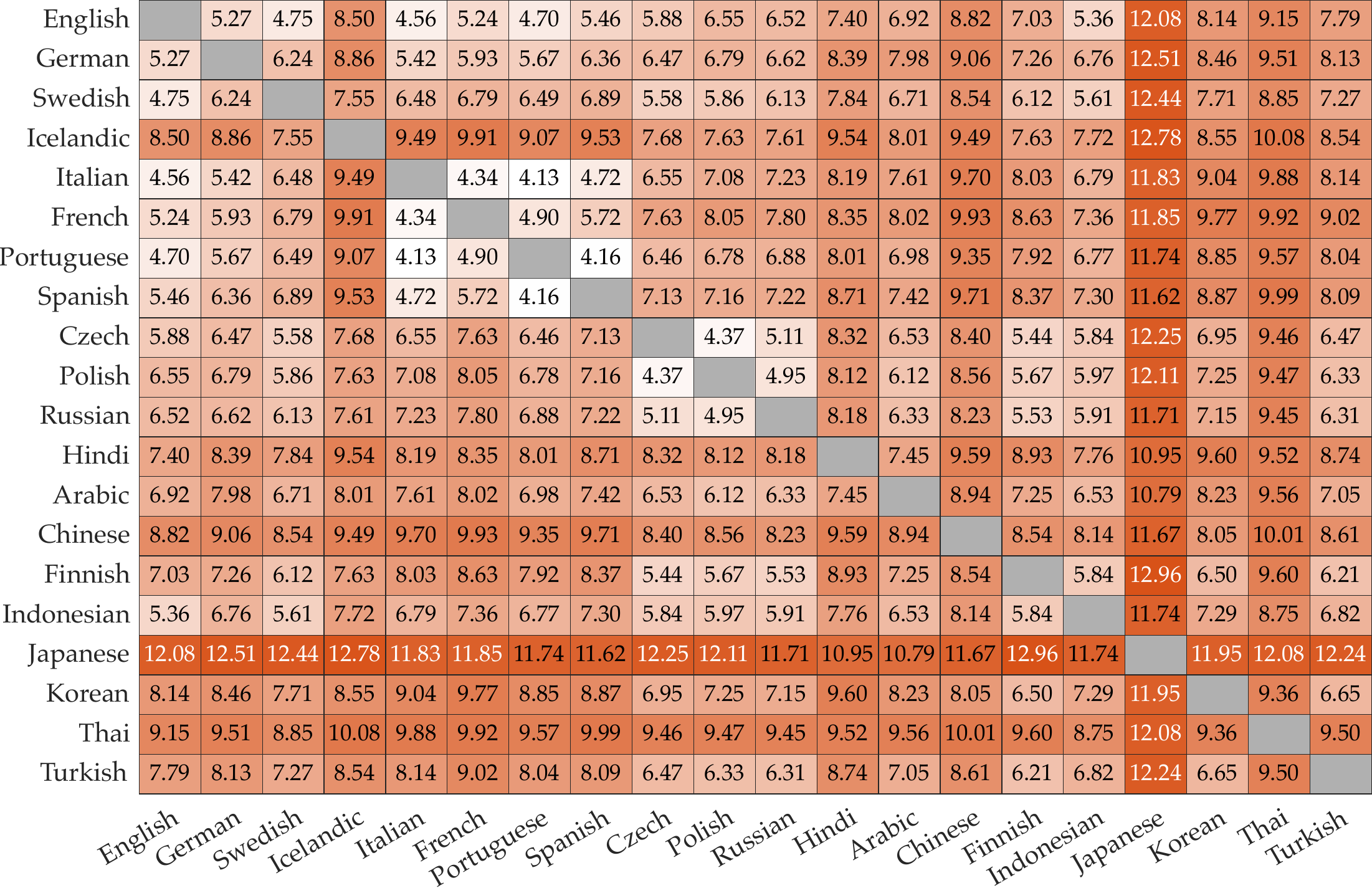}
\renewcommand{\figurename}{Supplementary Figure}
\caption{{\bf The language distance matrix of the SPA dataset.} The languages are ordered based on Glottolog 4.6 classification: Indo-European languages are listed first and grouped according to their subclasses (Germanic, Italic, Balto-Slavic and Indo-Iranian), and other languages are following in the alphabetical order.  }\label{sf8}
\end{center}
\end{figure}

\begin{table}[!ht]
    \centering
    \small
\begin{tabular}{ |p{1.7cm}||p{1.8cm}|p{1.8cm}|p{1.8cm}|p{1.8cm}|p{1.8cm}|p{1.8cm}| }
\hline
Closeness & 1st & 2nd & 3rd & 18th & 19th & 20th  \\
\hline
 \multicolumn{7}{|c|}{English} \\
 \hline
ENG & 4.28 (swe) & 5.06 (ger) & 5.08 (por) & 7.90 (chi) & 8.72 (tha) & 11.63 (jpn) \\
GER & 3.83 (por) & 4.07 (swe) & 4.15 (ita) & 7.53 (ice) & 7.79 (tha) & 9.23 (jpn) \\
FRE & 4.64 (spa) & 4.66 (ita) & 4.78 (por) & 7.86 (ice) & 8.92 (tha) & 10.72 (jpn) \\
ITA & 4.43 (por) & 4.81 (swe) & 4.94 (spa) & 8.26 (chi) & 9.16 (tha) & 11.13 (jpn) \\
SPA & 4.56 (ita) & 4.70 (por) & 4.75 (swe) & 8.82 (chi) & 9.15 (tha) & 12.08 (jpn) \\
 \hline
  \multicolumn{7}{|c|}{German} \\
 \hline
ENG & 5.06 (eng) & 5.61 (swe) & 5.91 (por) & 8.50 (hin) & 9.23 (tha) & 12.33 (jpn) \\
GER & 4.53 (eng) & 4.74 (swe) & 5.01 (por) & 7.57 (ice) & 7.91 (tha) & 9.80 (jpn) \\
FRE & 5.09 (eng) & 5.41 (swe) & 5.71 (ita) & 8.51 (chi) & 9.10 (tha) & 11.63 (jpn) \\
ITA & 5.35 (eng) & 5.59 (por) & 6.18 (spa) & 9.08 (chi) & 9.58 (tha) & 12.06 (jpn) \\
SPA & 5.27 (eng) & 5.42 (ita) & 5.67 (por) & 9.06 (chi) & 9.51 (tha) & 12.51 (jpn) \\
 \hline
  \multicolumn{7}{|c|}{Swedish} \\
 \hline
ENG & 4.28 (eng) & 5.19 (ind) & 5.61 (ger) & 7.93 (chi) & 8.44 (tha) & 12.26 (jpn) \\
GER & 4.07 (eng) & 4.32 (cze) & 4.59 (fin) & 7.15 (chi) & 7.24 (tha) & 9.75 (jpn) \\
FRE & 4.75 (ind) & 4.84 (eng) & 5.05 (cze) & 7.82 (chi) & 8.54 (tha) & 11.32 (jpn) \\
ITA & 4.81 (eng) & 5.79 (cze) & 5.88 (ind) & 8.47 (chi) & 8.91 (tha) & 11.98 (jpn) \\
SPA & 4.75 (eng) & 5.58 (cze) & 5.61 (ind) & 8.54 (chi) & 8.85 (tha) & 12.44 (jpn) \\
 \hline
  \multicolumn{7}{|c|}{Icelandic} \\
 \hline
ENG & 6.95 (swe) & 7.10 (fin) & 7.24 (ind) & 9.37 (hin) & 9.44 (tha) & 12.40 (jpn) \\
GER & 6.42 (cze) & 6.52 (fin) & 6.58 (rus) & 8.54 (tha) & 8.57 (fre) & 10.18 (jpn) \\
FRE & 7.12 (rus) & 7.31 (pol) & 7.32 (swe) & 9.26 (chi) & 9.71 (tha) & 11.84 (jpn) \\
ITA & 7.10 (fin) & 7.12 (rus) & 7.14 (cze) & 9.35 (hin) & 9.78 (tha) & 12.43 (jpn) \\
SPA & 7.55 (swe) & 7.61 (rus) & 7.63 (fin) & 9.91 (fre) & 10.08 (tha) & 12.78 (jpn) \\
 \hline

\end{tabular}
\renewcommand{\tablename}{Supplementary Table}
    \caption{{\bf The nearest and farthest languages to Germanic languages in language distance.} The 5 rows for each languages correspond to results in 5 datasets. Language are represented by their ISO 639-2/B codes.}
    \label{st1}
\end{table}

\begin{table}[ht!]
    \centering
    \small
\begin{tabular}{ |p{1.7cm}||p{1.8cm}|p{1.8cm}|p{1.8cm}|p{1.8cm}|p{1.8cm}|p{1.8cm}| }
\hline
Closeness & 1st & 2nd & 3rd & 18th & 19th & 20th  \\
\hline
 \multicolumn{7}{|c|}{Italian} \\
 \hline
ENG & 4.61 (por) & 5.18 (spa) & 5.49 (fre) & 9.33 (chi) & 9.67 (tha) & 11.72 (jpn) \\
GER & 3.44 (spa) & 3.53 (por) & 4.15 (eng) & 8.14 (tha) & 8.18 (chi) & 9.49 (jpn) \\
FRE & 4.39 (por) & 4.48 (spa) & 4.66 (eng) & 8.66 (ice) & 9.15 (tha) & 11.17 (jpn) \\
ITA & 4.63 (por) & 5.16 (spa) & 5.23 (eng) & 9.03 (chi) & 9.71 (tha) & 11.58 (jpn) \\
SPA & 4.13 (por) & 4.34 (fre) & 4.56 (eng) & 9.70 (chi) & 9.88 (tha) & 11.83 (jpn) \\
 \hline
  \multicolumn{7}{|c|}{French} \\
 \hline
ENG & 5.16 (por) & 5.49 (ita) & 5.80 (spa) & 9.44 (chi) & 9.83 (tha) & 11.92 (jpn) \\
GER & 4.46 (ita) & 4.51 (por) & 4.64 (spa) & 8.57 (ice) & 8.75 (tha) & 9.65 (jpn) \\
FRE & 5.13 (por) & 5.21 (ita) & 5.46 (spa) & 8.97 (chi) & 9.36 (tha) & 11.37 (jpn) \\
ITA & 4.74 (por) & 5.00 (eng) & 5.06 (spa) & 9.51 (chi) & 9.92 (tha) & 11.66 (jpn) \\
SPA & 4.34 (ita) & 4.90 (por) & 5.24 (eng) & 9.92 (tha) & 9.93 (chi) & 11.85 (jpn) \\
 \hline
  \multicolumn{7}{|c|}{Portuguese} \\
 \hline
ENG & 4.35 (spa) & 4.61 (ita) & 5.08 (eng) & 9.08 (chi) & 9.42 (tha) & 11.69 (jpn) \\
GER & 3.24 (spa) & 3.53 (ita) & 3.83 (eng) & 8.05 (ice) & 8.27 (tha) & 9.25 (jpn) \\
FRE & 3.60 (spa) & 4.39 (ita) & 4.78 (eng) & 8.56 (ice) & 9.27 (tha) & 10.65 (jpn) \\
ITA & 3.99 (spa) & 4.43 (eng) & 4.63 (ita) & 8.65 (chi) & 9.22 (tha) & 10.83 (jpn) \\
SPA & 4.13 (ita) & 4.16 (spa) & 4.70 (eng) & 9.35 (chi) & 9.57 (tha) & 11.74 (jpn) \\
 \hline
  \multicolumn{7}{|c|}{Spanish} \\
 \hline
ENG & 4.35 (por) & 5.18 (ita) & 5.72 (eng) & 9.12 (chi) & 9.37 (tha) & 11.55 (jpn) \\
GER & 3.24 (por) & 3.44 (ita) & 4.16 (eng) & 8.28 (tha) & 8.37 (ice) & 9.35 (jpn) \\
FRE & 3.60 (por) & 4.48 (ita) & 4.64 (eng) & 8.62 (ice) & 9.17 (tha) & 10.85 (jpn) \\
ITA & 3.99 (por) & 4.94 (eng) & 5.06 (fre) & 9.02 (chi) & 9.42 (tha) & 11.10 (jpn) \\
SPA & 4.16 (por) & 4.72 (ita) & 5.46 (eng) & 9.71 (chi) & 9.99 (tha) & 11.62 (jpn) \\
 \hline

\end{tabular}
\renewcommand{\tablename}{Supplementary Table}
    \caption{{\bf The nearest and farthest languages to Italic languages in language distance.} The 5 rows for each languages correspond to results in 5 datasets. Language are represented by their ISO 639-2/B codes.}
    \label{st2}
\end{table}

\begin{table}[ht!]
    \centering
    \small
\begin{tabular}{ |p{1.7cm}||p{1.8cm}|p{1.8cm}|p{1.8cm}|p{1.8cm}|p{1.8cm}|p{1.8cm}| }
\hline
Closeness & 1st & 2nd & 3rd & 18th & 19th & 20th  \\
\hline
 \multicolumn{7}{|c|}{Czech} \\
 \hline
ENG & 4.60 (pol) & 5.10 (rus) & 5.49 (fin) & 8.41 (hin) & 8.92 (tha) & 11.96 (jpn) \\
GER & 3.46 (pol) & 4.15 (rus) & 4.32 (fin) & 6.85 (chi) & 7.38 (tha) & 9.29 (jpn) \\
FRE & 4.56 (pol) & 4.82 (rus) & 5.05 (swe) & 8.13 (chi) & 8.70 (tha) & 11.08 (jpn) \\
ITA & 4.21 (pol) & 4.65 (rus) & 5.25 (fin) & 8.26 (hin) & 9.22 (tha) & 11.75 (jpn) \\
SPA & 4.37 (pol) & 5.11 (rus) & 5.44 (fin) & 8.40 (chi) & 9.46 (tha) & 12.25 (jpn) \\
 \hline
  \multicolumn{7}{|c|}{Polish} \\
 \hline
ENG & 4.60 (cze) & 4.79 (rus) & 5.58 (fin) & 8.21 (hin) & 8.80 (tha) & 11.75 (jpn) \\
GER & 3.46 (cze) & 3.89 (rus) & 4.55 (ara) & 6.93 (chi) & 7.46 (tha) & 9.16 (jpn) \\
FRE & 4.56 (rus) & 4.56 (cze) & 5.41 (ind) & 8.00 (chi) & 8.77 (tha) & 11.00 (jpn) \\
ITA & 4.21 (cze) & 4.39 (rus) & 5.53 (ara) & 8.25 (chi) & 9.15 (tha) & 11.40 (jpn) \\
SPA & 4.37 (cze) & 4.95 (rus) & 5.67 (fin) & 8.56 (chi) & 9.47 (tha) & 12.11 (jpn) \\
 \hline
  \multicolumn{7}{|c|}{Russian} \\
 \hline
ENG & 4.79 (pol) & 5.10 (cze) & 5.65 (ind) & 8.19 (hin) & 8.89 (tha) & 11.74 (jpn) \\
GER & 3.89 (pol) & 4.15 (cze) & 4.81 (ara) & 7.29 (chi) & 7.53 (tha) & 9.29 (jpn) \\
FRE & 4.14 (pol) & 4.82 (cze) & 5.14 (ind) & 7.72 (chi) & 8.72 (tha) & 10.70 (jpn) \\
ITA & 4.39 (pol) & 4.65 (cze) & 5.54 (ara) & 8.07 (hin) & 9.10 (tha) & 11.50 (jpn) \\
SPA & 4.95 (pol) & 5.11 (cze) & 5.53 (fin) & 8.23 (chi) & 9.45 (tha) & 11.71 (jpn) \\
 \hline
  \multicolumn{7}{|c|}{Hindi} \\
 \hline
ENG & 7.41 (eng) & 7.84 (swe) & 7.98 (ara) & 9.37 (ice) & 9.42 (tha) & 10.38 (jpn) \\
GER & 6.27 (eng) & 6.46 (ara) & 6.48 (por) & 8.26 (chi) & 8.28 (ice) & 8.33 (jpn) \\
FRE & 6.94 (eng) & 7.10 (rus) & 7.39 (spa) & 9.07 (ice) & 9.29 (tha) & 9.77 (jpn) \\
ITA & 7.74 (eng) & 7.80 (por) & 7.91 (ara) & 9.42 (chi) & 9.67 (tha) & 9.83 (jpn) \\
SPA & 7.40 (eng) & 7.45 (ara) & 7.76 (ind) & 9.59 (chi) & 9.60 (kor) & 10.95 (jpn) \\
 \hline

\end{tabular}
\renewcommand{\tablename}{Supplementary Table}
    \caption{{\bf The nearest and farthest languages to Balto-Slavic and Indo-Iranian languages in language distance.} The 5 rows for each languages correspond to results in 5 datasets. Language are represented by their ISO 639-2/B codes.}
    \label{st3}
\end{table}

\begin{table}[ht!]
    \centering
    \small
\begin{tabular}{ |p{1.7cm}||p{1.8cm}|p{1.8cm}|p{1.8cm}|p{1.8cm}|p{1.8cm}|p{1.8cm}| }
\hline
Closeness & 1st & 2nd & 3rd & 18th & 19th & 20th  \\
\hline
 \multicolumn{7}{|c|}{Arabic} \\
 \hline
ENG & 6.02 (rus) & 6.07 (pol) & 6.20 (ind) & 8.44 (chi) & 8.93 (tha) & 11.19 (jpn) \\
GER & 4.55 (pol) & 4.64 (cze) & 4.81 (rus) & 7.26 (chi) & 7.76 (tha) & 7.85 (jpn) \\
FRE & 5.33 (rus) & 5.75 (pol) & 6.08 (ind) & 8.51 (chi) & 9.05 (tha) & 10.59 (jpn) \\
ITA & 5.33 (pol) & 5.53 (rus) & 6.03 (ind) & 8.31 (chi) & 8.99 (tha) & 11.00 (jpn) \\
SPA & 6.12 (pol) & 6.33 (rus) & 6.53 (ind) & 8.94 (chi) & 9.56 (tha) & 10.79 (jpn) \\
 \hline
  \multicolumn{7}{|c|}{Chinese} \\
 \hline
ENG & 7.21 (kor) & 7.69 (ind) & 7.83 (fin) & 9.33 (ita) & 9.44 (fre) & 11.36 (jpn) \\
GER & 6.25 (kor) & 6.85 (cze) & 6.90 (ind) & 8.28 (ice) & 8.50 (fre) & 9.09 (jpn) \\
FRE & 7.50 (kor) & 7.62 (eng) & 7.67 (ind) & 8.97 (fre) & 9.26 (ice) & 10.09 (jpn) \\
ITA & 7.40 (kor) & 7.97 (rus) & 7.98 (ind) & 9.51 (fre) & 9.74 (tha) & 11.31 (jpn) \\
SPA & 8.05 (kor) & 8.14 (ind) & 8.23 (rus) & 9.93 (fre) & 10.01 (tha) & 11.67 (jpn) \\
 \hline
  \multicolumn{7}{|c|}{Finnish} \\
 \hline
ENG & 5.49 (cze) & 5.58 (pol) & 5.59 (ind) & 8.85 (hin) & 8.89 (tha) & 12.65 (jpn) \\
GER & 4.32 (cze) & 4.52 (tur) & 4.59 (swe) & 7.55 (tha) & 7.80 (fre) & 10.08 (jpn) \\
FRE & 5.42 (ind) & 5.51 (rus) & 5.68 (tur) & 8.27 (chi) & 9.13 (tha) & 12.22 (jpn) \\
ITA & 5.25 (cze) & 5.67 (rus) & 5.75 (tur) & 9.02 (hin) & 9.58 (tha) & 12.53 (jpn) \\
SPA & 5.44 (cze) & 5.53 (rus) & 5.67 (pol) & 8.93 (hin) & 9.60 (tha) & 12.96 (jpn) \\
 \hline
  \multicolumn{7}{|c|}{Indonesian} \\
 \hline
ENG & 5.19 (swe) & 5.24 (eng) & 5.59 (fin) & 8.19 (hin) & 8.39 (tha) & 11.77 (jpn) \\
GER & 4.73 (swe) & 4.78 (cze) & 4.83 (ara) & 6.98 (hin) & 7.34 (tha) & 8.96 (jpn) \\
FRE & 4.75 (swe) & 5.09 (eng) & 5.14 (rus) & 7.69 (hin) & 8.52 (tha) & 11.10 (jpn) \\
ITA & 5.74 (rus) & 5.83 (eng) & 5.85 (pol) & 8.15 (hin) & 8.81 (tha) & 11.31 (jpn) \\
SPA & 5.36 (eng) & 5.61 (swe) & 5.84 (fin) & 8.14 (chi) & 8.75 (tha) & 11.74 (jpn) \\
 \hline
 \multicolumn{7}{|c|}{Japanese} \\
 \hline
ENG & 10.38 (hin) & 11.19 (ara) & 11.36 (chi) & 12.33 (ger) & 12.40 (ice) & 12.65 (fin) \\
GER & 8.33 (hin) & 8.85 (ara) & 8.91 (kor) & 9.80 (ger) & 10.08 (fin) & 10.18 (ice) \\
FRE & 9.77 (hin) & 10.09 (chi) & 10.59 (ara) & 11.63 (ger) & 11.84 (ice) & 12.22 (fin) \\
ITA & 9.83 (hin) & 10.83 (por) & 11.00 (ara) & 12.06 (ger) & 12.43 (ice) & 12.53 (fin) \\
SPA & 10.79 (ara) & 10.95 (hin) & 11.62 (spa) & 12.51 (ger) & 12.78 (ice) & 12.96 (fin) \\
 \hline
  \multicolumn{7}{|c|}{Korean} \\
 \hline
ENG & 6.13 (fin) & 6.25 (tur) & 6.75 (ind) & 8.96 (hin) & 9.11 (fre) & 11.52 (jpn) \\
GER & 5.21 (fin) & 5.49 (tur) & 5.92 (ind) & 7.72 (hin) & 8.10 (fre) & 8.91 (jpn) \\
FRE & 5.95 (fin) & 5.98 (tur) & 6.37 (swe) & 8.35 (tha) & 8.52 (hin) & 10.85 (jpn) \\
ITA & 6.26 (fin) & 6.30 (tur) & 6.87 (rus) & 9.06 (fre) & 9.16 (hin) & 11.61 (jpn) \\
SPA & 6.50 (fin) & 6.65 (tur) & 6.95 (cze) & 9.60 (hin) & 9.77 (fre) & 11.95 (jpn) \\
 \hline
  \multicolumn{7}{|c|}{Thai} \\
 \hline
ENG & 8.39 (ind) & 8.44 (swe) & 8.45 (kor) & 9.67 (ita) & 9.83 (fre) & 12.07 (jpn) \\
GER & 7.24 (swe) & 7.34 (ind) & 7.38 (cze) & 8.54 (ice) & 8.75 (fre) & 9.65 (jpn) \\
FRE & 8.35 (kor) & 8.52 (ind) & 8.54 (swe) & 9.36 (fre) & 9.71 (ice) & 10.97 (jpn) \\
ITA & 8.81 (ind) & 8.91 (swe) & 8.96 (kor) & 9.78 (ice) & 9.92 (fre) & 11.87 (jpn) \\
SPA & 8.75 (ind) & 8.85 (swe) & 9.15 (eng) & 10.01 (chi) & 10.08 (ice) & 12.08 (jpn) \\
 \hline
  \multicolumn{7}{|c|}{Turkish} \\
 \hline
ENG & 5.64 (fin) & 6.03 (ind) & 6.13 (pol) & 8.59 (hin) & 8.82 (tha) & 11.90 (jpn) \\
GER & 4.52 (fin) & 4.82 (cze) & 4.95 (pol) & 7.44 (fre) & 7.60 (tha) & 9.33 (jpn) \\
FRE & 5.68 (fin) & 5.84 (pol) & 5.96 (cze) & 8.30 (chi) & 8.84 (tha) & 11.07 (jpn) \\
ITA & 5.75 (fin) & 6.07 (rus) & 6.21 (pol) & 8.75 (hin) & 9.50 (tha) & 11.99 (jpn) \\
SPA & 6.21 (fin) & 6.31 (rus) & 6.33 (pol) & 9.02 (fre) & 9.50 (tha) & 12.24 (jpn) \\
 \hline

\end{tabular}
\renewcommand{\tablename}{Supplementary Table}
    \caption{{\bf The nearest and farthest languages to non-Indo-European languages in language distance.} The 5 rows for each languages correspond to results in 5 datasets. Language are represented by their ISO 639-2/B codes.}
    \label{st4}
\end{table}

\begin{figure}[ht]
\begin{center}
\includegraphics[scale=0.6825]{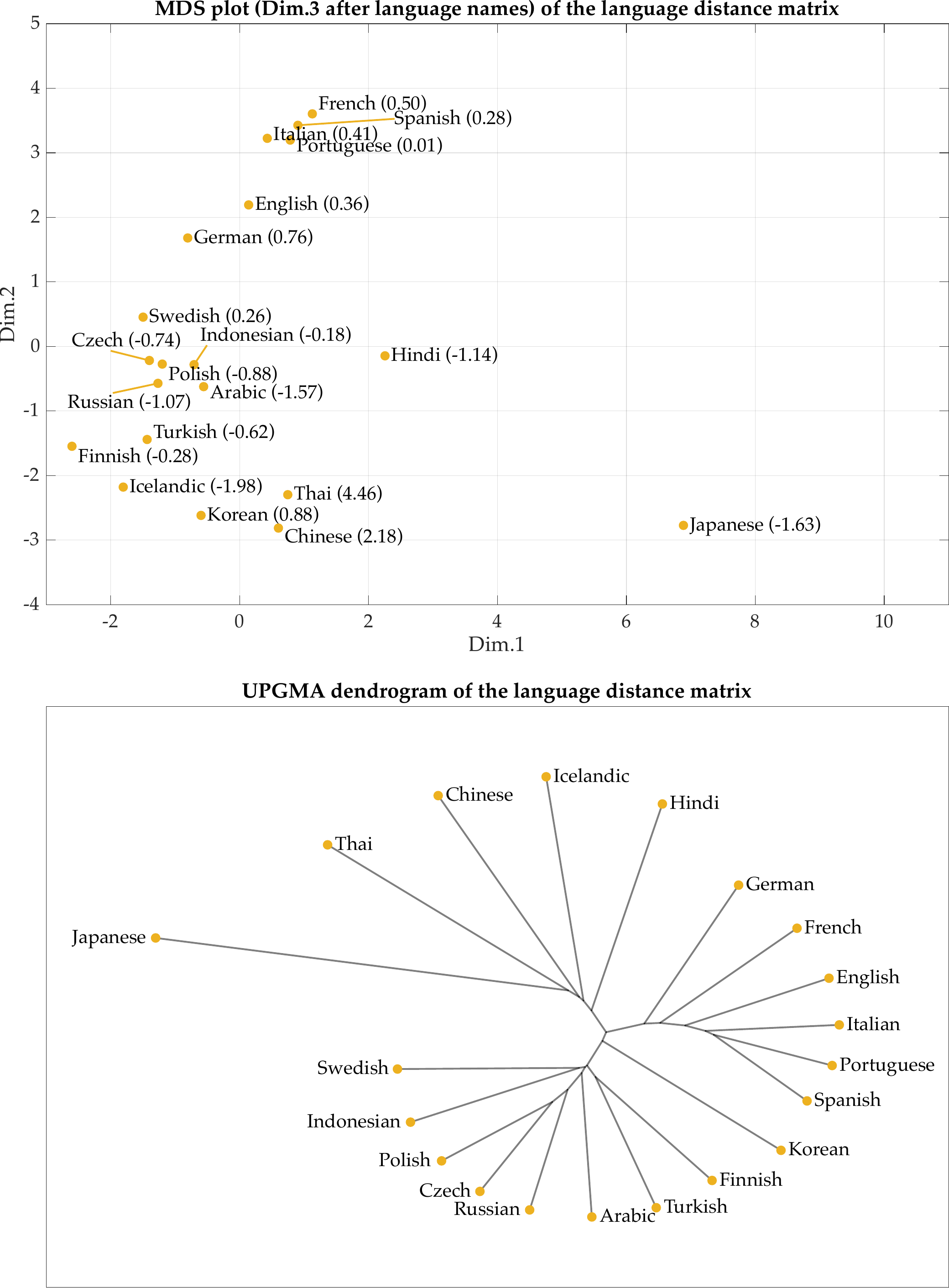}
\renewcommand{\figurename}{Supplementary Figure}
\caption{{\bf Visualizations the language distance matrix of the GER dataset.} Top: the multidimensional scaling plot of the language distance matrix of the GER dataset. Bottom: the UPGMA dendrogram constructed based on the language distance matrix of the GER dataset.}\label{sf9}
\end{center}
\end{figure}

\begin{figure}[ht]
\begin{center}
\includegraphics[scale=0.6825]{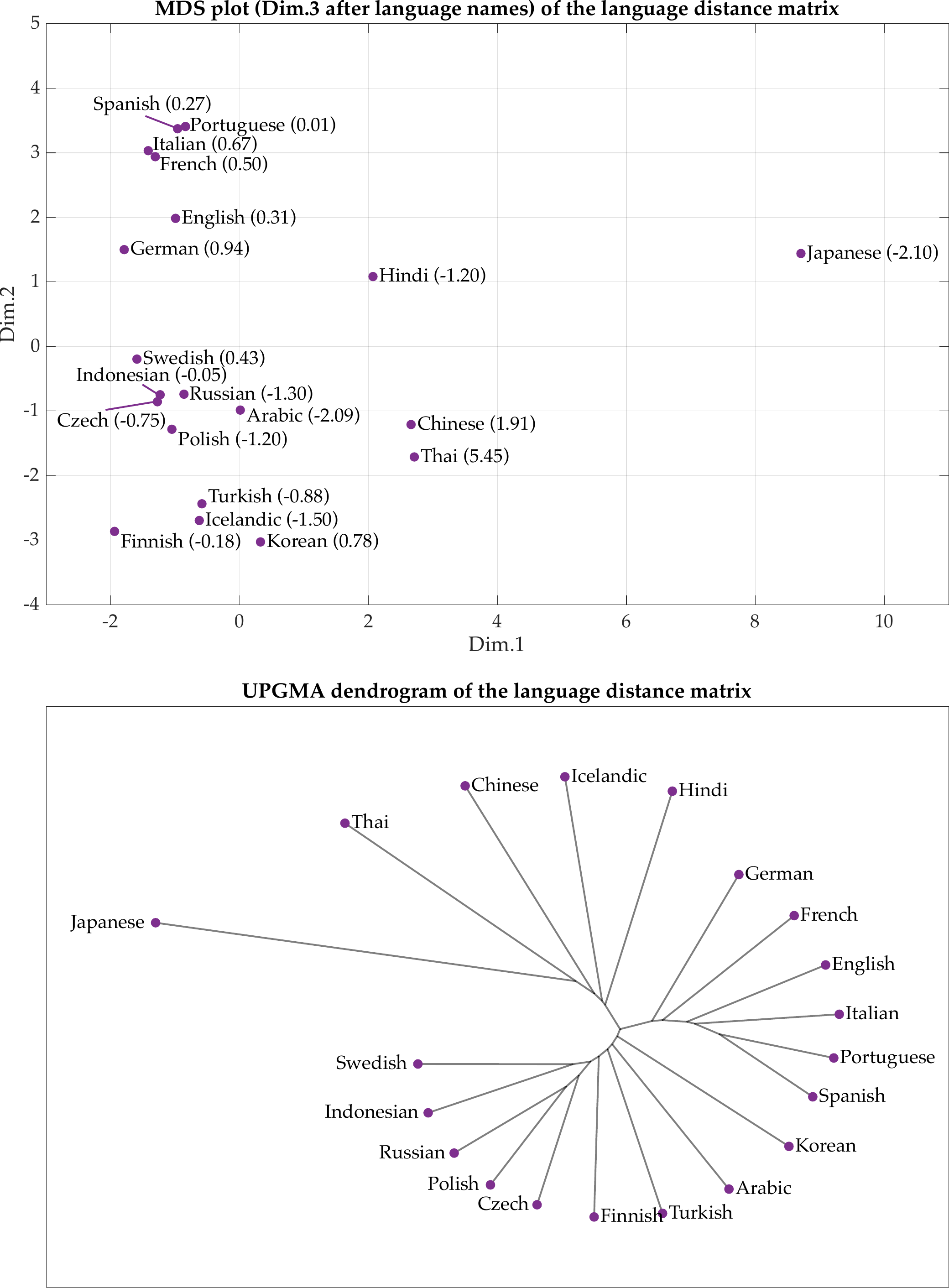}
\renewcommand{\figurename}{Supplementary Figure}
\caption{{\bf Visualizations the language distance matrix of the FRE dataset.} Top: the multidimensional scaling plot of the language distance matrix of the FRE dataset. Bottom: the UPGMA dendrogram constructed based on the language distance matrix of the FRE dataset.}\label{sf10}
\end{center}
\end{figure}

\begin{figure}[ht]
\begin{center}
\includegraphics[scale=0.6825]{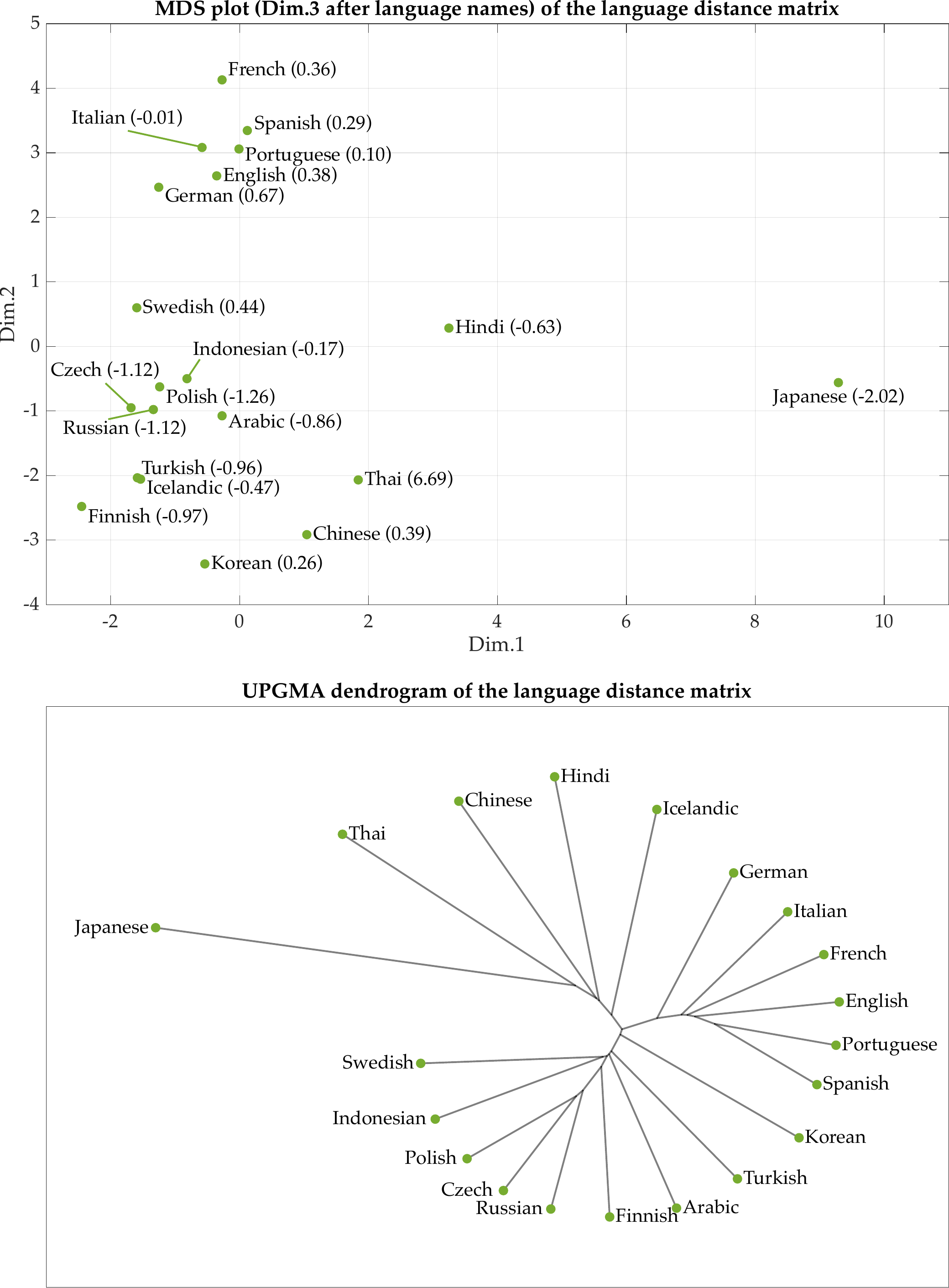}
\renewcommand{\figurename}{Supplementary Figure}
\caption{{\bf Visualizations the language distance matrix of the ITA dataset.} Top: the multidimensional scaling plot of the language distance matrix of the ITA dataset. Bottom: the UPGMA dendrogram constructed based on the language distance matrix of the ITA dataset. }\label{sf11}
\end{center}
\end{figure}

\begin{figure}[ht]
\begin{center}
\includegraphics[scale=0.6825]{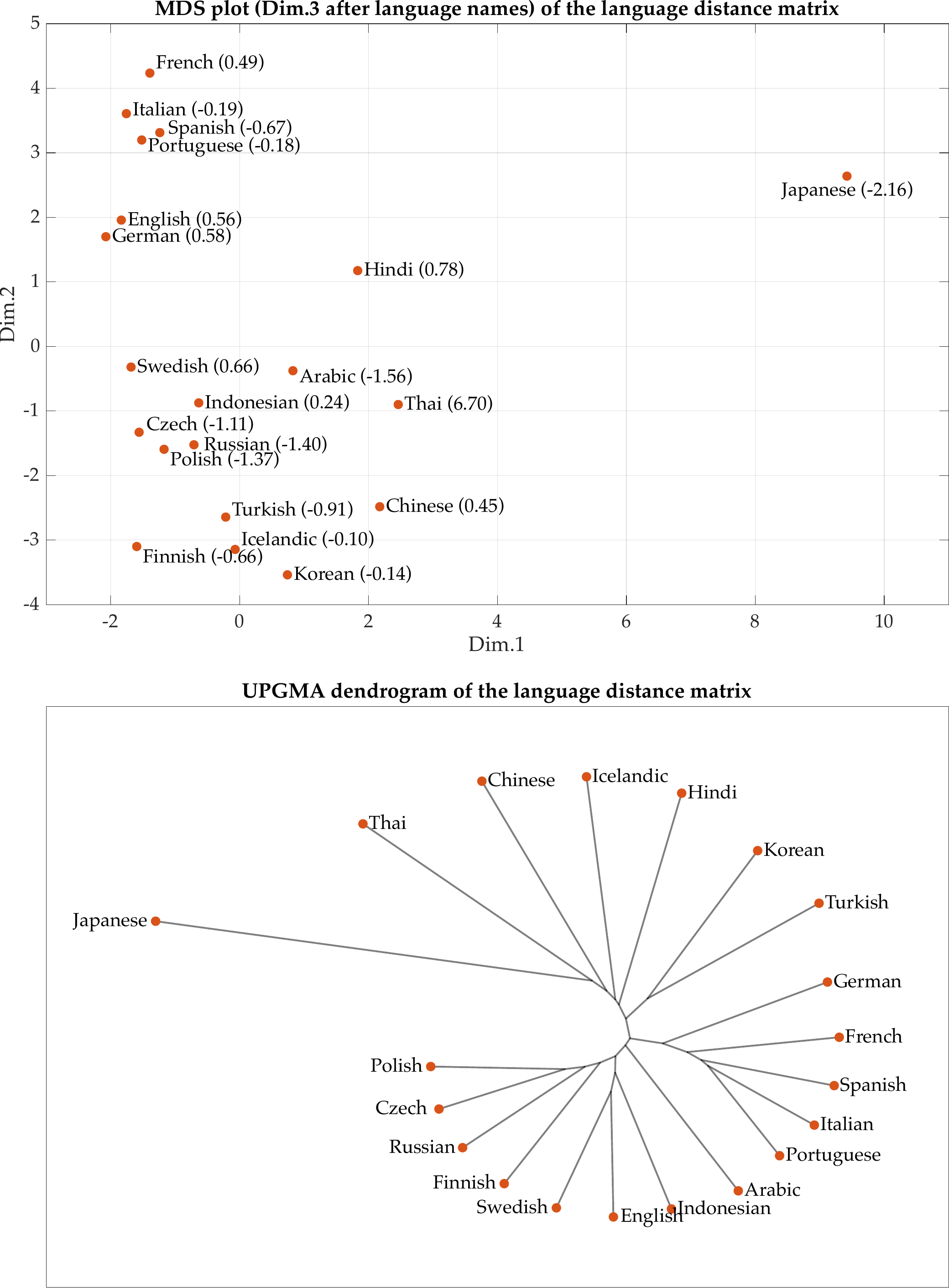}
\renewcommand{\figurename}{Supplementary Figure}
\caption{{\bf Visualizations the language distance matrix of the SPA dataset.} Top: the multidimensional scaling plot of the language distance matrix of the SPA dataset. Bottom: the UPGMA dendrogram constructed based on the language distance matrix of the SPA dataset. }\label{sf12}
\end{center}
\end{figure}

\begin{figure}[!ht]
\begin{center}
\includegraphics[scale=0.6825]{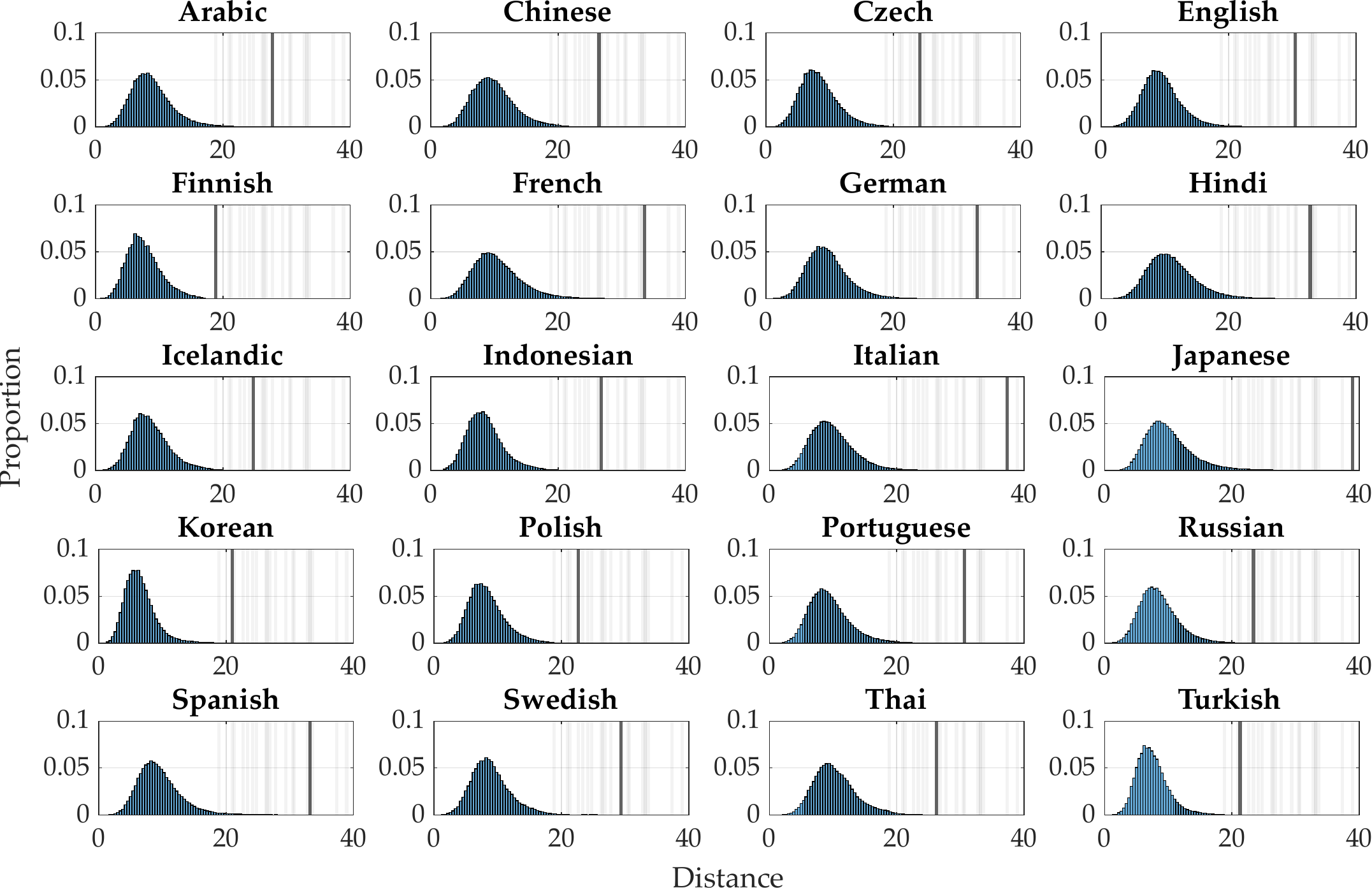}
\renewcommand{\figurename}{Supplementary Figure}
\caption{{\bf The diameters and the distributions of pairwise sentence distances in 20 corpora of the ENG dataset.} 
The diameters of 20 languages are showed as vertical lines in the panels: the solid line for the present corpus and transparent lines for other 19 corpora.  }\label{sf13}
\end{center}
\end{figure}

\begin{figure}[!ht]
\begin{center}
\includegraphics[scale=0.6825]{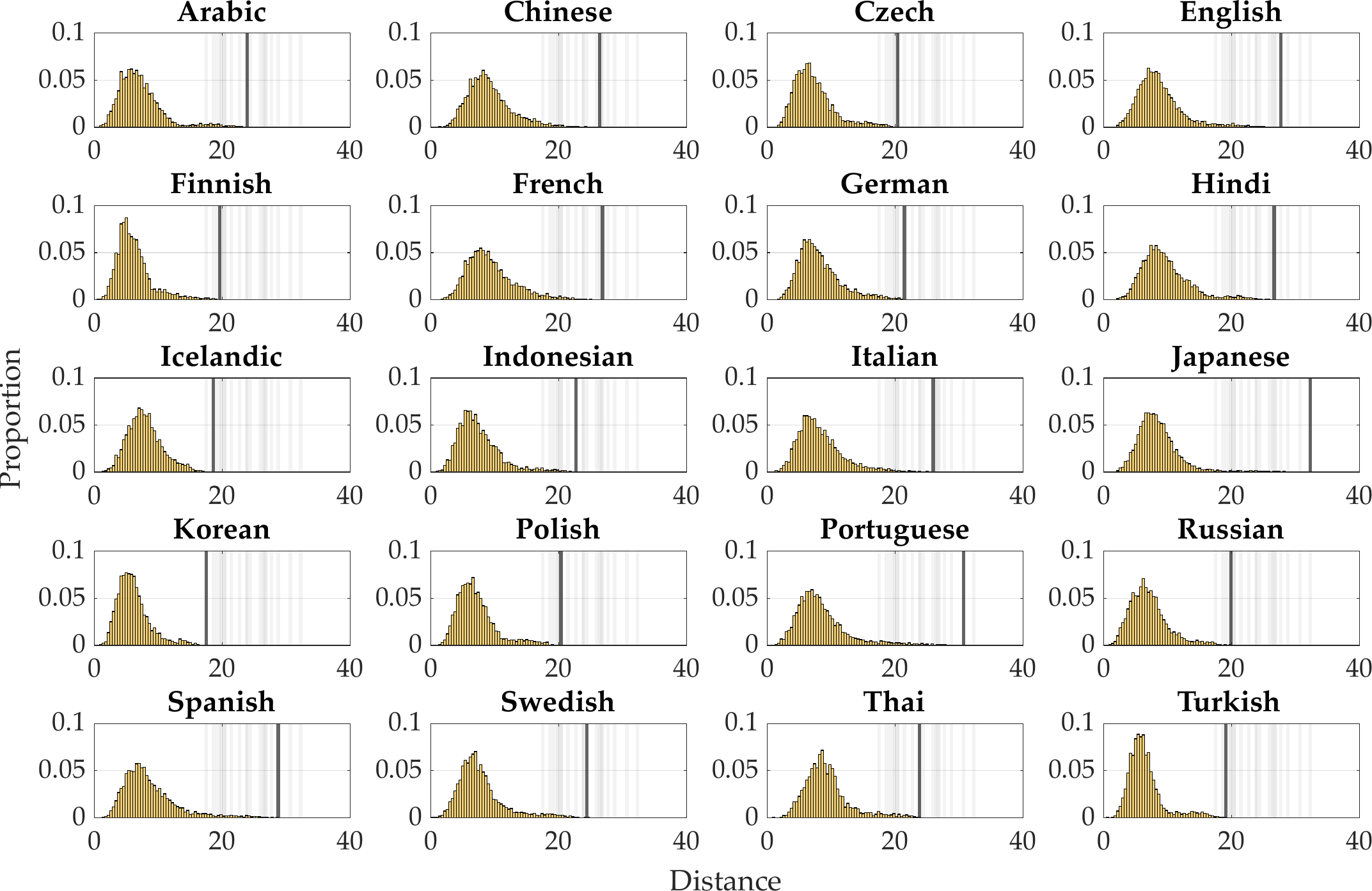}
\renewcommand{\figurename}{Supplementary Figure}
\caption{{\bf The diameters and the distributions of pairwise sentence distances in 20 corpora of the GER dataset.} 
The diameters of 20 languages are showed as vertical lines in the panels: the solid line for the present corpus and transparent lines for other 19 corpora. }\label{sf14}
\end{center}
\end{figure}

\begin{figure}[ht]
\begin{center}
\includegraphics[scale=0.6825]{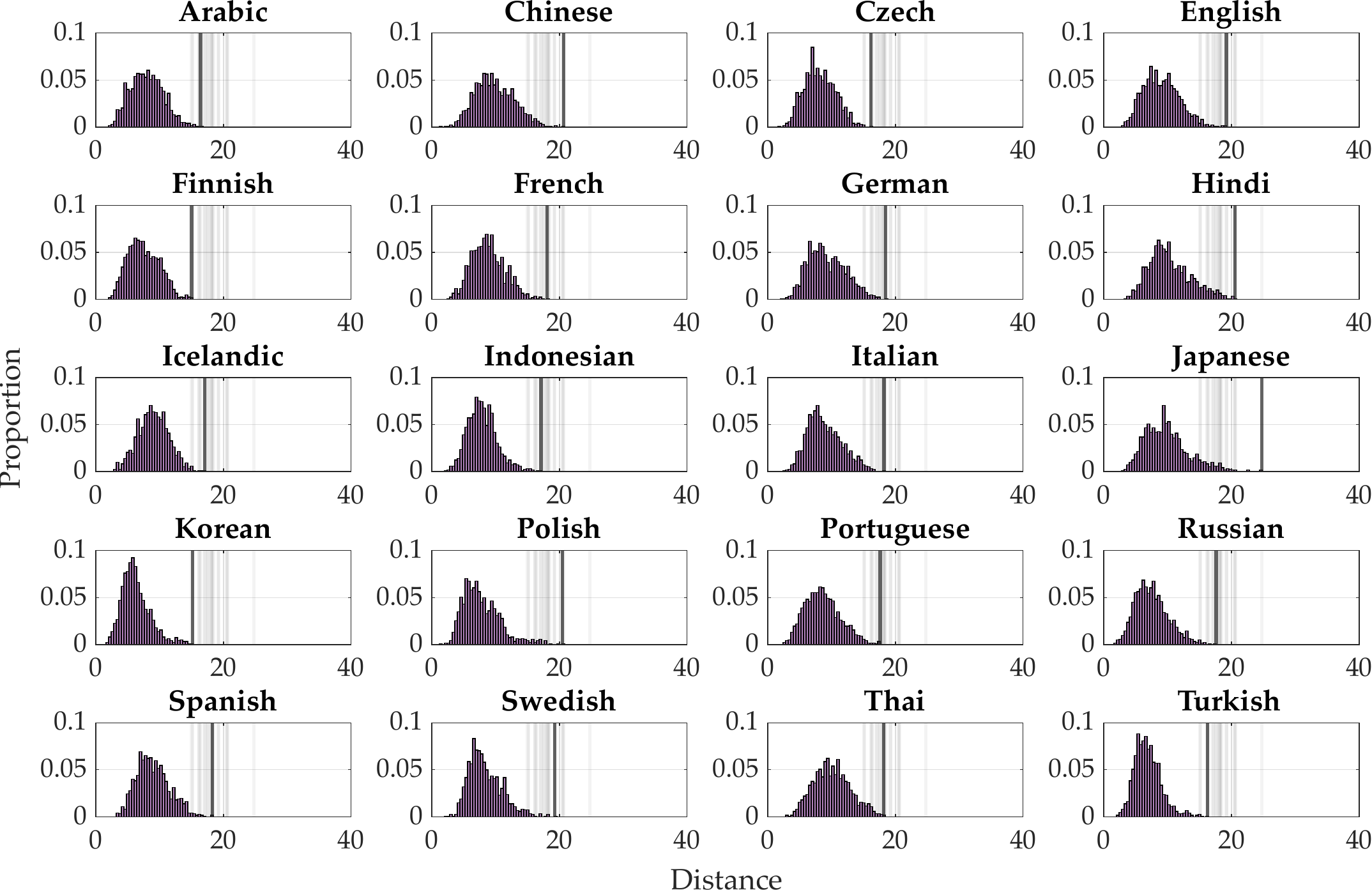}
\renewcommand{\figurename}{Supplementary Figure}
\caption{{\bf The diameters and the distributions of pairwise sentence distances in 20 corpora of the FRE dataset.} 
The diameters of 20 languages are showed as vertical lines in the panels: the solid line for the present corpus and transparent lines for other 19 corpora.  }\label{sf15}
\end{center}
\end{figure}

\begin{figure}[ht]
\begin{center}
\includegraphics[scale=0.6825]{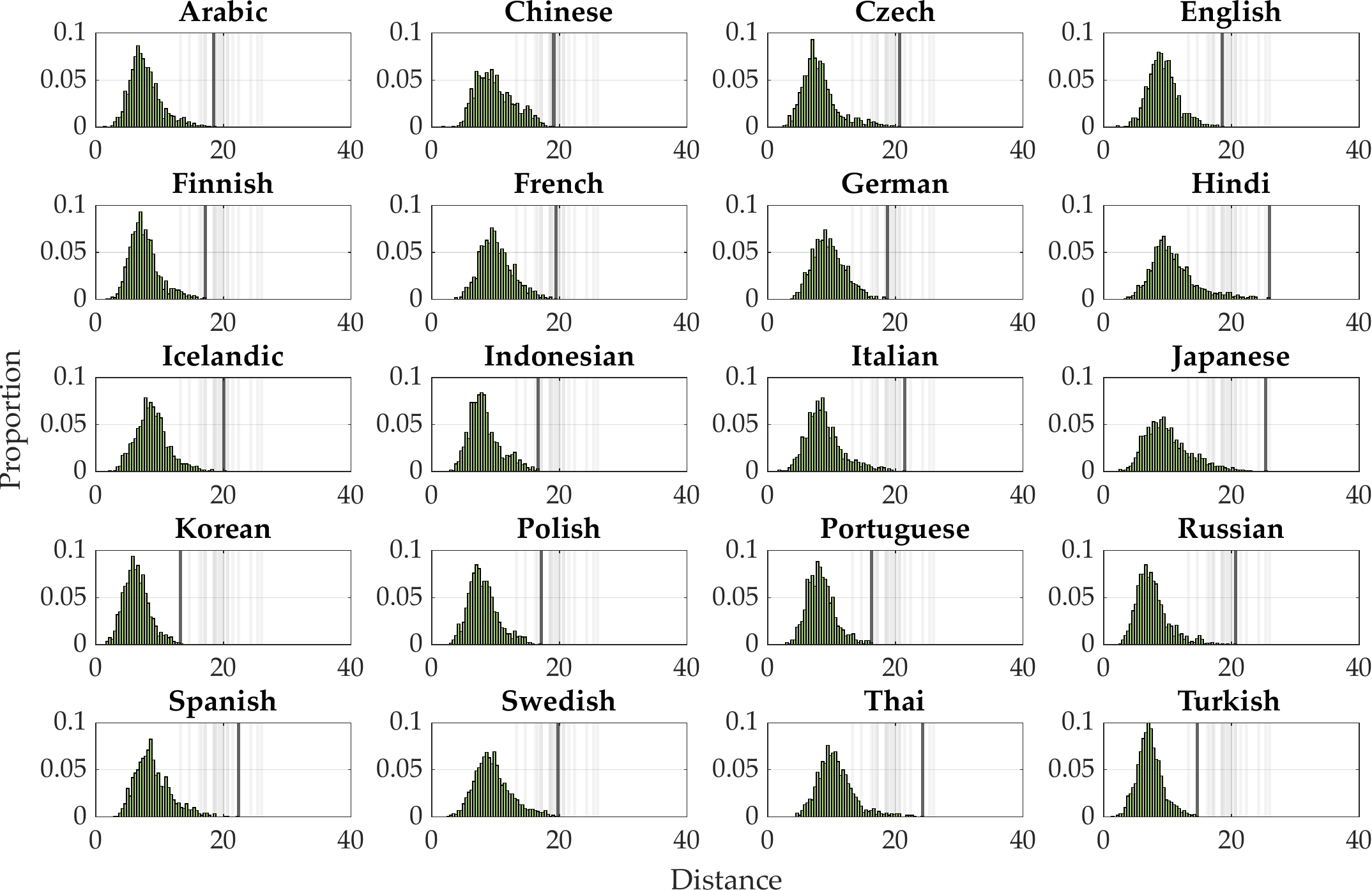}
\renewcommand{\figurename}{Supplementary Figure}
\caption{{\bf The diameters and the distributions of pairwise sentence distances in 20 corpora of the ITA dataset.} 
The diameters of 20 languages are showed as vertical lines in the panels: the solid line for the present corpus and transparent lines for other 19 corpora.  }\label{sf16}
\end{center}
\end{figure}

\begin{figure}[ht]
\begin{center}
\includegraphics[scale=0.6825]{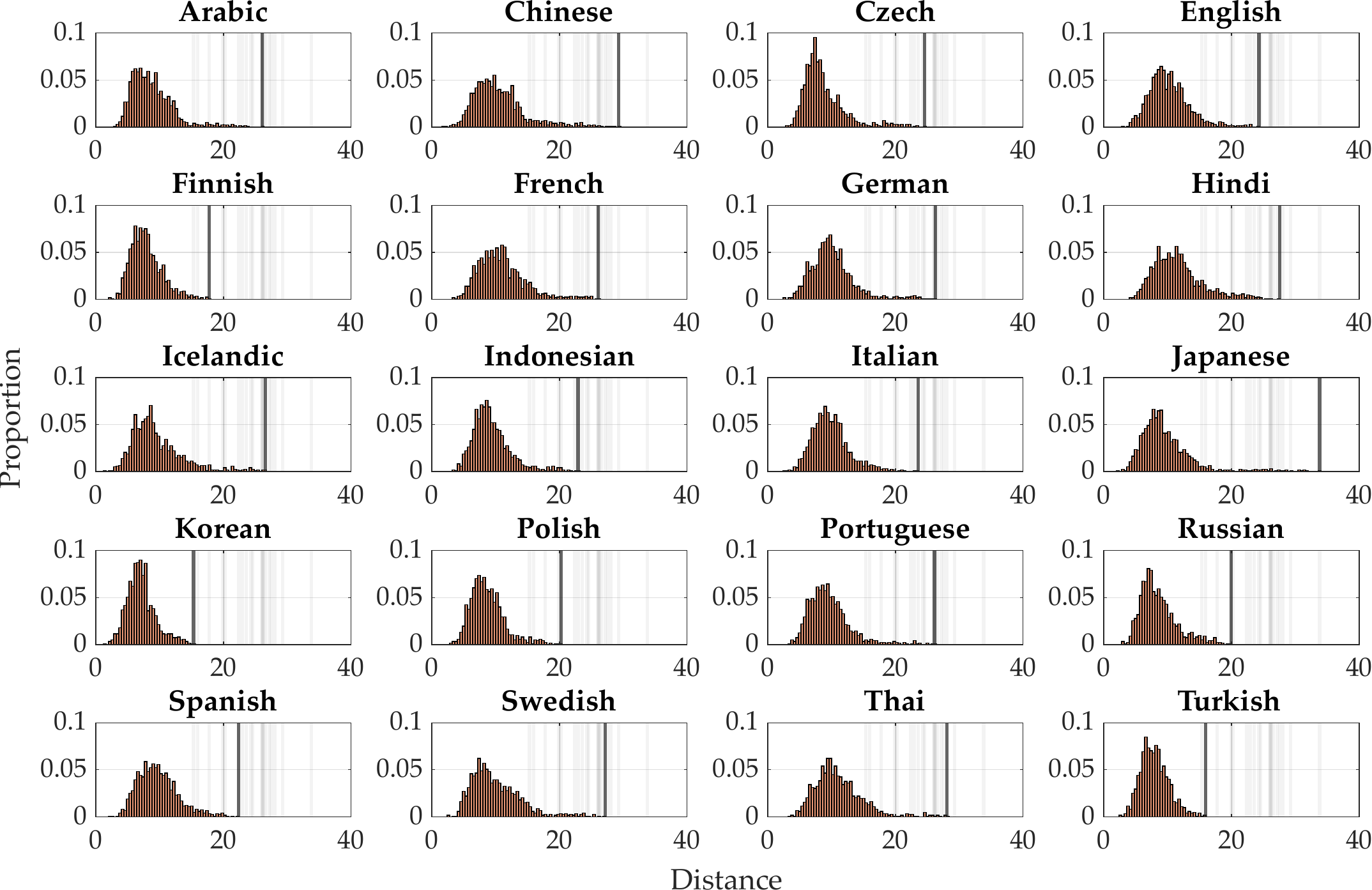}
\renewcommand{\figurename}{Supplementary Figure}
\caption{{\bf The diameters and the distributions of pairwise sentence distances in 20 corpora of the SPA dataset.} 
The diameters of 20 languages are showed as vertical lines in the panels: the solid line for the present corpus and transparent lines for other 19 corpora.  }\label{sf17}
\end{center}
\end{figure}

\end{document}